\documentclass[sn-mathphys-num]{sn-jnl}

\usepackage{graphicx}%
\usepackage{multirow}%
\usepackage{amsmath,amssymb,amsfonts}%
\usepackage{amsthm}%
\usepackage{mathrsfs}%
\usepackage{hyperref}%
\usepackage{nicefrac}%
\usepackage{natbib}%
\usepackage[title]{appendix}%
\usepackage{xcolor}%
\usepackage{textcomp}%
\usepackage{manyfoot}%
\usepackage{booktabs}%
\usepackage{algorithm}%
\usepackage{algorithmicx}%
\usepackage{algpseudocode}%
\usepackage{listings}%
\usepackage{cleveref}%

\usepackage{xcolor}
\usepackage[utf8]{inputenc} 
\usepackage[T1]{fontenc}    
\usepackage{hyperref}       
\usepackage{natbib}
\usepackage{boldline} 
\usepackage{url}            
\usepackage{booktabs}       
\usepackage{amsfonts}       
\usepackage{nicefrac}       
\usepackage{microtype}      
\usepackage{lipsum}
\usepackage{fancyhdr}
\usepackage{colortbl}
\usepackage{makecell}
\usepackage{enumitem}
\usepackage{xcolor}
\usepackage{dsfont}
\usepackage{graphicx}       
\usepackage{algorithm}
\usepackage{algpseudocode}
\usepackage{amsmath}
\usepackage{cleveref}
\usepackage{subcaption}

\theoremstyle{thmstyleone}%
\theoremstyle{thmstyletwo}%

\theoremstyle{thmstylethree}%

\begin{document}

\title[Article Title]{EnTri: Ensemble learning with tri-level representations for explainable scene recognition}


\author*[1]{\fnm{Amirhossein} \sur{Aminimehr}}\email{amir\_aminimehr@comp.iust.ac.ir}

\author[1]{\fnm{Amirali} \sur{Molaei}}\email{amirali\_molaei@comp.iust.ac.ir}

\author[2]{\fnm{Erik} \sur{Cambria}}\email{cambria@ntu.edu.sg}

\affil[1]{\orgdiv{School of Computer Engineering}, \orgname{Iran University of Science and Technology}, \orgaddress{\street{Hengam St., Resalat Square}, \city{Tehran}, \postcode{13114-16846}, \state{Tehran}, \country{Iran}}}

\affil[2]{\orgdiv{School of Computer Science and Engineering}, \orgname{Nanyang Technological University}, \orgaddress{\street{50 Nanyang Avenue}, \city{Singapore}, \postcode{639798}, \state{Singapore}, \country{Singapore}}}


\abstract{Scene recognition based on deep-learning has made significant progress, but there are still limitations in its performance due to challenges posed by inter-class similarities and intra-class dissimilarities. Furthermore, prior research has primarily focused on improving classification accuracy, yet it has given less attention to achieving interpretable, precise scene classification. Therefore, we are motivated to propose EnTri, an ensemble scene recognition framework that employs ensemble learning using a hierarchy of visual features. EnTri represents features at three distinct levels of detail: pixel-level, semantic segmentation-level, and object class and frequency level. By incorporating distinct feature encoding schemes of differing complexity and leveraging ensemble strategies, our approach aims to improve classification accuracy while enhancing transparency and interpretability via visual and textual explanations. To achieve interpretability, we devised an extension algorithm that generates both visual and textual explanations highlighting various properties of a given scene that contribute to the final prediction of its category. This includes information about objects, statistics, spatial layout, and textural details. Through experiments on benchmark scene classification datasets, EnTri has demonstrated superiority in terms of recognition accuracy, achieving competitive performance compared to state-of-the-art approaches, with an accuracy of 87.69\%, 75.56\%, and 99.17\% on the MIT67, SUN397, and UIUC8 datasets, respectively.}




\maketitle

\section{Introduction}\label{sec1}

Scene recognition is a fundamental computer vision task that aims to categorize scene images among predetermined classes, serving as a foundation for a wide range of applications such as robotics, autonomous driving, and remote sensing. A scene refers to a place where human beings can perform actions and navigate based on their perceptions of the environment~\citep{xiao2010sun}. Recognizing scenes requires discerning the ambient contents of the image, including the background environment and the layout of objects in the scene, and then assembling the semantic information of these elements into an effective representation that describes the scene.
	
\begin{figure*}
    \centering	\includegraphics[width=0.6\textwidth]{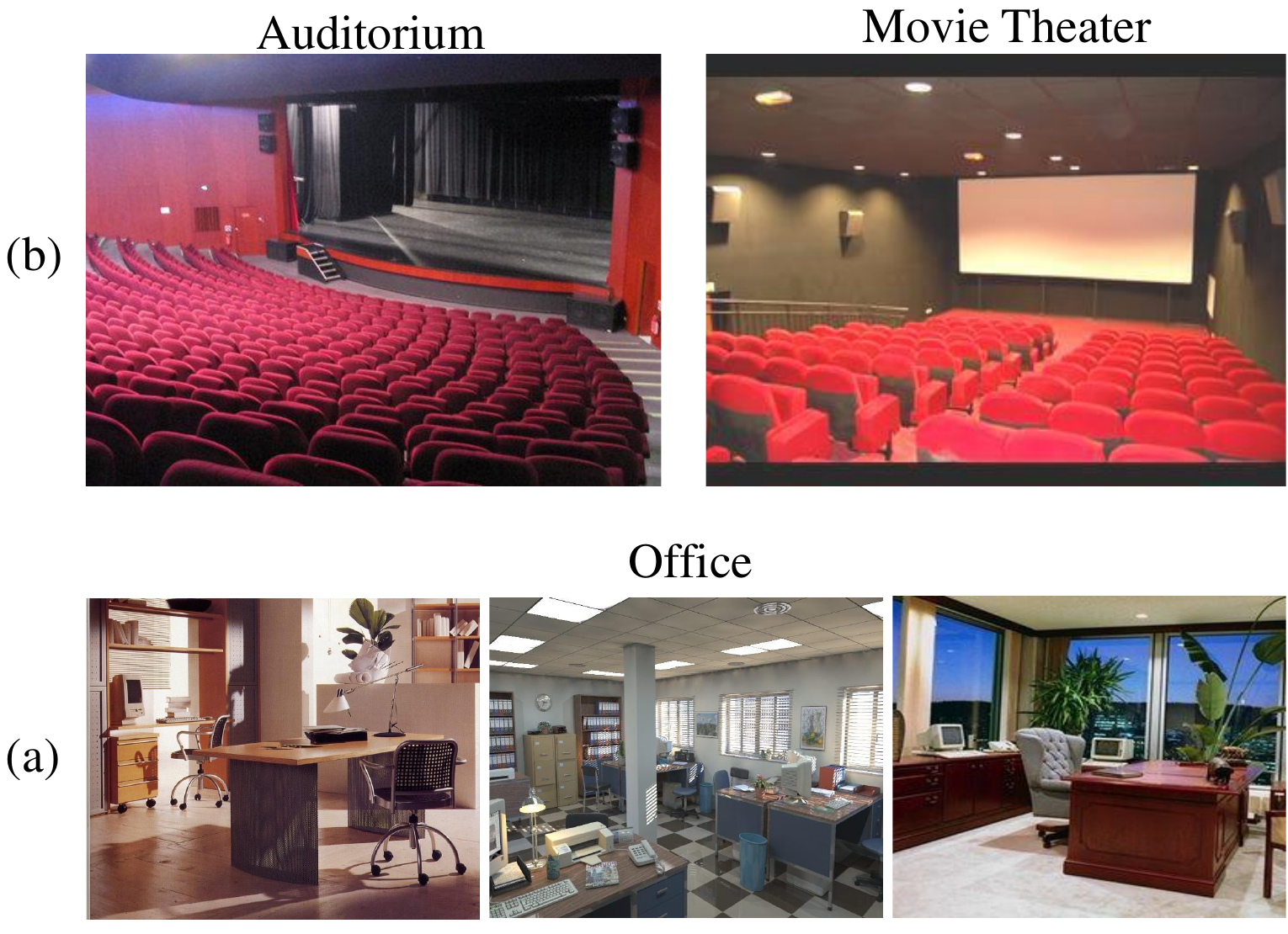}
    \caption{Demonstrations of inter-class similarity and intra-class variation. a) Images from the auditorium and movie theater classes have a high degree of similarity (inter-class similarity). b) Images of the office demonstrate a considerable degree of intra-class diversity, suggesting a wide spectrum of visual features within the category.}
    \label{challenges}
\end{figure*}

Scene recognition is challenging, mainly due to inter-class similarities and intra-class variations~\citep{li2020text,wang2021robust,sitaula2021content,fan2022indoor}. For instance, auditoriums and movie theaters share similar visual appearances (though they have different objects such as a screen or a curtain, as shown in~\Cref{challenges}(a)), and offices have variations across different scenes (though they have similar objects such as computers, chairs, or tables, as shown in~\Cref{challenges}(b)), which may confuse the recognition system. Furthermore, constructing a scene representation that captures crucial semantic information reflecting the complexity of the data is challenging, particularly when dealing with a large number of categories.

Research in scene recognition frameworks can be categorized into ones that are based on hand-crafted engineering and methods that employ automatic feature extraction without human intervention. Hand-engineered features are based on manually designing and selecting features utilizing different techniques to capture spatial characteristics, local features, object-based concepts, and holistic representations of scenes~\citep{oliva2001modeling,lowe2004distinctive,csurka2004visual,li2010object}. However, hand-crafted features require notable domain expertise and significant human effort, resulting in inefficiency. Consequently, Deep Convolutional Neural Networks (DCNNs) have largely replaced them due to their superior representation learning ability~\citep{voulodimos2018deep,ragusa2019survey,machado2021adversarial}. They have demonstrated that they attain superior classification performance when trained on extensive datasets.
After AlexNet's~\citep{krizhevsky2017imagenet} introduction, deep convolutional neural networks, such as VGG~\citep{simonyan2014very}, Inception~\citep{szegedy2015going}, ResNet~\citep{he2016deep}, and DenseNet~\citep{huang2017densely}, impressively advanced the field of image classification due to their great ability of capturing locally correlated image values~\citep{lecun2015deep, he2019bag, li2021survey} and learning features that enhance efficiency compared to hand-crafted methods. Utilizing deep convolutional neural networks can enhance scene recognition performance, but it remains challenging due to the intricate spatial layout, intra-class variation, and inter-class similarity, which weaken discriminability among scenes~\citep{cheng2018scene,xiong2020msn}. In addition, due to the rise of extensive scene-centric datasets, a simple CNN-generated representation model is inadequate to accurately discriminate large-scale scenes~\cite{lin2022scene}. As a result, research has shifted its focus to developing more representative features by incorporating contextual information, such as objects, or proposing strategies to effectively extract features that facilitate decision-making boundaries. Furthermore, several studies have employed ensemble learning strategies to leverage the complementary strengths of multiple feature levels or recognition models~\citep{nanni2013heterogeneous,bai2018softly,hernandez2019indoor}.

Besides improving accuracy, recognition systems based on deep neural networks face the challenge of ambiguity in the reasoning behind the model's predictions~\citep{ribeiro2016should, dovsilovic2018explainable, islam2021explainable}. This boils down to the black-box nature of deep neural networks, which induces a lack of trust and ethical concerns regarding the model's decisions, especially in high-stakes applications such as medical analysis and self-driving vehicles~\citep{miller2019explanation,das2020opportunities,zhang2022explainable}. Along this direction, numerous efforts have been made to help understand how a model solves a problem and makes decisions. Typically, an explanation should help us get the answers to the following questions: Why did the model predict this category? Is the prediction of the model reliable? Which parts of the input led to this output? For vision tasks, researchers usually provide visual explanations, including saliency maps and heatmaps, to highlight important components in the input that contribute to the prediction~\citep{tjoa2020survey,islam2021explainable}. In general, a good visual explanation manifests the fine-grained and highly class-discriminative information of the input image~\citep{selvaraju2017grad}. While heatmaps can effectively convey information to users, their effectiveness may be limited for non-expert users who may have difficulty comprehending the correlation between input features and visual explanations. The interpretation of heatmaps by users can also be more uncertain compared to texts. On the other hand, texts are generally easier for non-expert users to comprehend than heatmaps, which may be attributed to the fragmented and dispersed nature of heatmap-identified areas within the original image.
	
\begin{figure*}
    \centering	\includegraphics[width=0.6\textwidth]{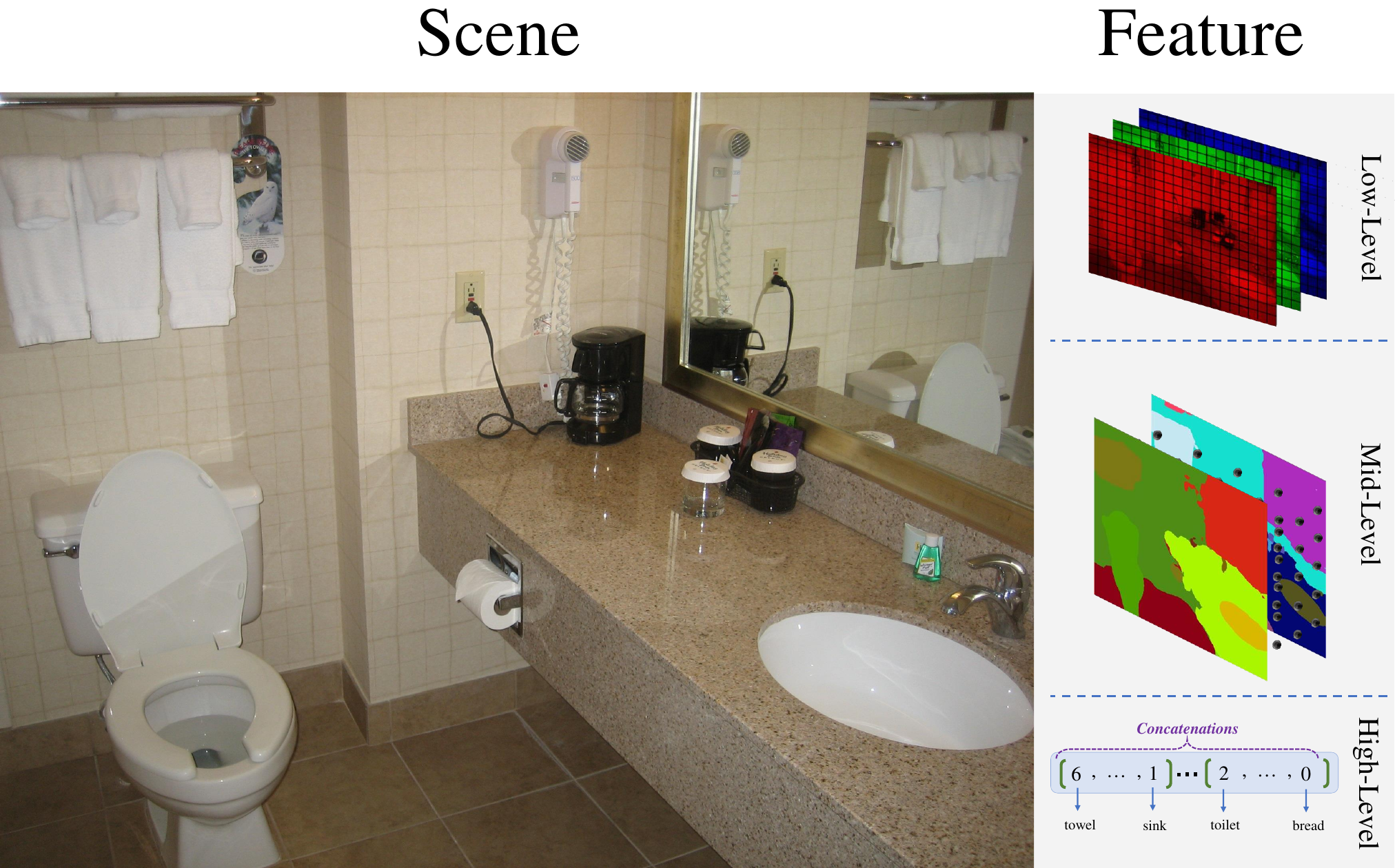}
    \caption{An example showing different levels of representation extracted from a scene using EnTri.}
    \label{levels}
\end{figure*}

Towards such challenges, we propose EnTri, an ensemble scene recognition framework that employs the ensemble learning technique using a hierarchy of visual features. We utilize pre-trained deep networks to build these features, employ an ensemble of classifiers to produce classification scores, and then combine these predictions as input for another model, which is a fully-connected network that generates the final category of the scene. Inspired by various works in visual recognition~\cite{yang2015multi,lin2017feature}, we formulate three levels of perceptual features as follows: 1) low-level features, representing the entire image itself, which include the textural information or intensity levels of every pixel; 2) mid-level features, built using semantic segmentation, which capture the spatial and categorical attributes of objects in the image; and 3) high-level features, extracted using object detectors, which abstractly represent the objects and their associated statistics (object frequency in the scene).~\Cref{levels} provides an illustrative example of different levels of representation. By integrating semantic features at three distinct levels, our method produces richer semantic features, resulting in competitive accuracy when compared to state-of-the-art methods. Moreover, EnTri incorporates networks that are pre-trained on various datasets for building mid-level and high-level representations, making it applicable to scene datasets that lack semantic segmentation or object detection annotations. To ensure interpretability, we further design an extension algorithm that generates both visual and textual explanations. These explanations highlight the scene properties that contributed to the final prediction of the scene category, such as object categories and frequencies, object locations, and textural information, as well as confidence scores. Textual explanations are advantageous in overcoming the limitations of heatmaps, which may suffer from non-uniformity and dispersion in relevant areas. With textual explanations, the user can identify specific items and regions that were considered during the prediction process by reading. Therefore, to enhance interpretability and facilitate a deeper understanding of the prediction process, we provide both visual and textual explanations. 

\vspace{10pt}

In particular, our contributions are as follows:

\begin{itemize}
    \item We introduce EnTri, a framework for learning scene recognition that integrates three distinct levels of feature representation—low, mid, and high—in an ensemble learning fashion. It incorporates three models, each leveraging a particular feature level to make predictions, then combines them using a weighted combination strategy, which assigns weights based on their performance on the validation set. The resulting combination is then passed as input to a fully-connected network to determine the final scene category.

    \item We propose an algorithm that utilizes visual and textual explanations to elucidate the prediction of scene recognition by highlighting key attributes such as object categories, frequencies, locations, and textural information, as well as confidence scores.

     \item EnTri's modular design allows the feature-specific analysis of low, mid, and high-level scene attributes that impact recognition performance while also facilitating the generation of both visual and textual explanations, making it easier to diagnose and optimize system behavior. 

    \item Extensive experiments are conducted on three benchmark scene datasets, including MIT67~\citep{quattoni2009recognizing}, Sun397~\citep{xiao2010sun}, and UIUC8~\citep{li2007and}. The results demonstrate the competitive performance of EnTri with the existing state-of-the-art methods.
\end{itemize}

\vspace{10pt}

The rest of the paper is structured as follows: Section 2, provides an overview of the work related to our method. Section 3 presents a table that defines the notations used throughout the paper. In Section 4, the principles of the proposed EnTri is introduced in detail. Section 5 provides experimental analysis and outcomes. Finally, Section 6 concludes this work and identifies various potential areas for future research.

\section{Related work}\label{sec2} 

This section covers the relevant studies that share common elements and ideas with our work. We overview the literature in scene recognition, covering a range of techniques based on either hand-engineered or deep-learning extracted features, as well as ensemble learning. We also explore the research made in explainable AI for image classification, which can include the integration of visual explanations, textual explanations, or both.

\subsection{Scene recognition}

Based on feature extraction as a vital stage of scene recognition, research in this area can be divided into two main categories of traditional and modern approaches. Additionally, we explore ensemble recognition frameworks, which employ the same learning strategy as ours.
\subsubsection{Traditional approaches}
Here, we briefly review the literature on scene recognition models, starting from the earliest research with manual extraction of features. To effectively predict scenes, visual descriptors have been broadly utilized to represent the textural and structural features of the scenes. To encode spatial characteristics, GIST~\cite{oliva2001modeling} utilizes the Fourier transform and principal components analysis to create perceptual dimensions concerning spatial layouts. SIFT~\citep{lowe2004distinctive} extracts scale-invariant local features called keypoints of objects, then matches properties of new scenes to these feature vectors based on euclidean distance. Inspired by the Bag-of-Words method, Bag-of-Visual-Words (BoVW)~\cite{csurka2004visual} treats visual information as words and utilizes visual descriptors (such as SIFT) to represent scenes as disordered arrays of visual features. Since BoVW disregards spatial features, Lazebnik et al. extend it and propose spatial pyramid matching (SPM)~\cite{lazebnik2006beyond}, which works by iteratively dividing the image into increasingly fine sub-regions and computing histograms of local features found inside each section. CENTRIST~\cite{wu2010centrist} builds holistic representations that capture structural properties of the scenes by using a non-parametric local transform to model local structures. To increase discriminative power, Object Bank (OB)~\cite{li2010object} emphasized the significance of semantically more complex features associated with the contents of the scene (such as objects) and proposed high-level object-based representations using the responses of multiple object detectors. Nonetheless, the curse of dimensionality impedes its practicability.

\subsubsection{Modern approaches}
    \label{modern approaches}
 The aforementioned methods still lack sufficient discriminative power for complex scenes and are inefficient as they require considerable domain expertise and human effort. The ability of deep convolutional neural networks (DCNNs) to learn and extract features automatically regardless of the image type facilitates the construction of more effective holistic image representations for scene recognition. They are capable of exploiting high-level perceptual semantics, with the small receptive field of earlier layers capturing edges and textures and building up to higher-level features and bigger receptive fields in the last layers to represent abstract concepts, such as object shapes and structural properties~\cite{lin2017feature,qian2018learning}. Consequently, most researchers include CNN models as the backbone of their work. Following the introduction of AlexNet~\citep{krizhevsky2017imagenet}, subsequent architectures, such as VGG~\citep{simonyan2014very}, Inception~\citep{szegedy2015going}, ResNet~\citep{he2016deep}, DenseNet~\citep{huang2017densely}, and ConvNeXt~\cite{liu2022convnet} have consistently advanced the field. However, due to intra-class variations and inter-class similarity, coupled with the rise of extensive scene-centric datasets, a simple CNN-generated representation model is insufficient to accurately discriminate complex scenes. Thus, the research community has shifted its attention towards developing more representative features by incorporating contextual information, such as objects, or employing multi-scale hierarchies. Additionally, there are other miscellaneous methods for promoting efficient feature representations.

Multi-scale approaches focus on extracting features in a hierarchy of multiple stages, starting from high-resolution to low-resolution. One notable method that addresses this is the multi-scale orderless pooling (MOP-CNN) introduced by Gong et al.~\cite{gong2014multi}. MOP-CNN aggregates features extracted from local patches at different scales to construct a more invariant, orderless representation. Likewise, DAG-CNN~\cite{yang2015multi} framework configures CNNs in a directed acyclic graph (DAG) style to effectively capture multiple levels of features extracted from the CNN layers. Herranz et al.~\citep{herranz2016scene} utilizes a multi-scale design that combines multiple CNNs processing the image patches, increasing in size with each scale in parallel. 

In terms of object-based and contextual scene recognition, LG-CNN~\cite{xie2017lg} computed the gradients of feature maps via a pre-trained CNN to detect part candidates and approximates global object locations using saliency detectors as the input to a CNN model designed with local parts and global discrimination. Wang et al. address the CNN issues regarding its data-hungry and black-box nature and aggregate descriptors of local scene patches based on a weakly supervised semantic codebook to construct a hybrid visual representation called VSAD~\cite{wang2017weakly}. Cheng et al.~\cite{cheng2018scene} proposed to incorporate object co-occurrence information by utilizing object score vectors to obtain their distribution in local patches. They identified discriminative objects based on the scene's posterior probability, acquired from co-occurrence patterns. However, this method ignores the spatial information associated with objects. To address this, López-Cifuentes et al.~\cite{lopez2020semantic} used semantic segmentation in an attention module based on an end-to-end CNN to generate a semantic-weighted representation, followed by a gating process to incorporate the contextual information to enhance accuracy. Chen et al.~\citep{chen2019scene} proposed to represent each object in the image as word embeddings by a scene parser and utilized them to refine the ordered top@5 scores of the classification model (ResNet~\citep{he2016deep}) to enhance accuracy. It performs well but doesn't address the importance of speed in visual data processing for interacting robots. Finally, Heikel et al.~\citep{heikel2022indoor} utilized the YOLO~\citep{redmon2016you} object detection model to detect objects in indoor environments, represented them as TF-IDF vector semantics, and used them as features to predict room categories through a random forest classifier.

Other techniques aim to build representative features by exploring alternative concepts and approaches. For instance, RLML-LSR~\citep{wang2021robust} employs local metric learning approach for scene recognition that integrates a discriminative metric function and least squares regression regularization to enhance robustness. Likewise, EML-ELM~\citep{wang2022embedding} combines metric learning and extreme learning machines (ELM) to generate discriminative features and develop a classifier with faster learning, respectively. Additionally, AdaNFF~\citep{zou2022adanff} is a recognition framework based on feature fusion, combining unprocessed features extracted from images to reduce redundancy and provide complementary elements in an end-to-end process. Overall, modern research has incorporated various techniques, including semantic segmentation, extreme learning machines, metric learning, and feature fusion, to enhance the accuracy of scene recognition frameworks.

\subsubsection{Ensemble approaches}	
Ensemble learning has been used for many years in research for scene recognition and image classification to improve the accuracy and robustness of models. By combining the predictions of multiple models, ensemble learning addresses the limitations of individual models and enhances the overall performance of the system. In ensemble with traditional machine learning models, Nanni et al. utilized a variety of texture descriptors and combined them with an ensemble of support vector machines for classification~\citep{nanni2013heterogeneous}. Furthermore, in ensemble learning using CNN models, Soft Combination~\citep{bai2018softly} leverages CNN's hierarchical structure to train an ensemble of classifiers on features extracted from different layers, then combines these classifiers using a strategy that employs a set of weights for combination. Similarly, Hernandez et al.~\citep{hernandez2019indoor} implemented a weighted voting scheme to combine an ensemble of classifiers for more robust scene recognition. 

Compared to all the methods discussed thus far, in this paper we attempt to ameliorate the shortcoming of the prior research by leveraging the contextual cues of the scene through the application of three levels of features. These features are used to enhance the recognition performance, employing the stacking ensemble method that combines the predictions generated from these features in a weighted fashion. To prevent overfitting and mitigate potential biases, predictions at different feature levels are generated using multiple models with different architectures.


\subsection{Explainable image classification}
Different methods have been proposed to make a machine learning model interpretable. These methodologies can be broadly categorized into two classes: model-agnostic approaches, such as Lime~\citep{ribeiro2016should} SHAP~\citep{lundberg2017unified}, and Layer-wise Relevance Propagation (LRP)~\cite{bach2015pixel} which are adaptable to different types of ML models; and architecture-specific methods, such as Grad-CAM~\citep{selvaraju2017grad}, which is developed for image classification convolutional neural networks (CNNs). In image classification, an explanation clarifies the properties and components influencing the prediction of the scene category. To this end, in saliency-based methods, important areas of the input image are highlighted based on their relevance to the model's prediction using saliency maps, which are matrices with the same dimensions as the input image. For instance, Lime and SHAP explain individual predictions by perturbing the pixels of the input and observing their effect on the output, and Grad-CAM is a technique that generates a localization map highlighting relevant parts of an image for predicting a target concept by using gradients of the concept's logits through the final convolutional layer, resulting in importance weights for each neuron. Likewise, LRP propagates the relevance value of the nodes in the network's output layer back through the layers to the input image, assigning relevance scores to each pixel based on its contribution to the final output~\cite{bach2015pixel}.

As aforementioned, textual explanations ameliorate intelligibility in terms of recognizing significant objects compared to visual explanations. Many tasks, such as image captioning, have attempted to generate textual information about an image. Our research shares similarities to this task, which aims to generate textual narratives describing an image through an encoder-decoder architecture~\citep{stefanini2022show}. Over the years, numerous studies have focused on developing image captioning systems that effectively convey visual information in a textual format~\citep{vinyals2015show,huang2019attention}. However, these methods are limited in their ability to provide textual explanations for the model's prediction, as they rely on labels to learn how to generate descriptive captions. Conversely, the following methods actually generate textual explanations for the prediction of scene recognition systems. For instance, Anjomshoae et al.~\citep{anjomshoae2021context} divides the scene image into its segments and use partial masking to assess the contextual importance of each segment. It then generates explanations based on feature importance, which are presented visually as a saliency map highlighting the components with the highest influence on the prediction, along with a textual explanation listing the features and their level of effect on the output. It also provides a color bar graph showing the importance value of each feature. Additionally, Aminimehr et al. proposed TbExplain~\citep{aminimehr4385953tbexplain}, which takes a different approach by utilizing XAI techniques and an object detector to explain scene classification models. The system generates saliency heatmaps using an XAI method, and identifies influential objects by measuring the area of the overlapped region between the heatmap and the bounding box of each object detected by the object detector. Textual explanations are then provided by integrating these influential objects with pre-defined natural language text. Compared to these methods, we provide textual explanations from the viewpoint of three levels of features extracted from the scene. Moreover, these textual explanations present the degree of agreement between the predictions made using the three features through ensemble learning and the final prediction of our framework. Our textual explanations not only highlight the critical attributes such as object categories, frequencies, locations, and textural cues but also provide scores that represent the contribution of individual objects towards the prediction. We also provide visual explanations along with textual to enhance interpretability.

	


	\section{Notation} 
	To ensure clarity and avoid confusion for readers,~\Cref{tab:notations} presents a comprehensive list of the notations used throughout the paper, along with their corresponding definitions.

\begingroup
\def\arraystretch{1.4}%
\begin{table}[h]
	\centering
        \fontsize{7.5}{8.75}\selectfont
	\caption{Summary of Notations}
		\begin{tabular}{|c|m{4.3cm} |c|m{4.3cm}|}
			\hline
			\rowcolor{gray!30}
			\textbf{Notation} & \textbf{Definition} & \textbf{Notation} & \textbf{Definition} \\ \hline
			\rowcolor{blue!5}
			$x$ & An input scene & $HLF$ &  High-level feature representation of the scene $x$   \\
			\rowcolor{blue!10}
			$M_L$ & Number of low-level discriminator models  & $Y_L$ & Output matrix of the low-level sub-model \\
			\rowcolor{blue!5}
			$M_M$ & Number of mid-level discriminator models  & $Y_M$ & Output matrix of mid-level sub-model \\
			\rowcolor{blue!10}
			$M_H$ & Number of high-level discriminator models  & $Y_H$ & Output matrix of high-level sub-model \\
			\rowcolor{blue!5}
			$N_{PM_{s}}$ & Number of pre-trained segmentation models  & $\alpha$ & The exponent of the weight values for prediction combination \\
			\rowcolor{blue!10}
			$N_{PM_{o}}$ & Number of pre-trained object detection models  & $F_L$ & A set of low-level discriminator models  \\
			\rowcolor{blue!5}
			$N_C$ & Number of scene categories  & $f_L^{(m)}$ & the $m^{th}$ Low-level discriminator model \\
			\rowcolor{blue!10}
			$N_x$ & Dimension of the scene image along the x-axis & $F_M$ & Set of mid-level discriminator models \\
			\rowcolor{blue!5}
			$N_y$ & Dimension of the scene image along the y-axis  & $f_M^{(m)}$ & the $m^{th}$ Mid-level discriminator model \\
			\rowcolor{blue!10}
			$SV$ & Output softmax vector of a model  & $F_H$ & Set of high-level discriminator models  \\
			\rowcolor{blue!5}
			$PM_s$ & Set of pre-trained segmentation models  & $f_H^{(m)}$ & The $m^{th}$ High-level discriminator model \\
			\rowcolor{blue!10}
			$pm_{s}^{(n)}$ & The $n^{th}$ Pre-trained segmentation model  & $OV$ & Set of object vectors generated from object detection models\\
			\rowcolor{blue!5}
			$PM_o$ & Set of pre-trained object detection models  & $ov_{n}$ &  The $n^{th}$ A set of high-level discriminator models \\
			\rowcolor{blue!10}
			$pm_{o}^{(n)}$ & The $n^{th}$ pre-trained object detection model  & $SM$ & Set of segmentation maps generated from semantic segmentation models\\
			\rowcolor{blue!5}
			$MLF$ & Mid-level feature representation of the scene $x$  & $sm_{n}$ & Segmentation map of the $n^{th}$ model \\ \hline

		\end{tabular}
	
\label{tab:notations}
\end{table}
\endgroup

	\section{Approach}
	This paper proposes EnTri to remedy the problem associated with scene classification by using an explainable hierarchical ensemble framework that encodes features at different levels and enables better comprehension of the process's underlying logic. Our framework includes several sub-models and a meta-model following the principles and concepts of the stacked generalization ensemble technique~\cite{wolpert1992stacked}. Since we aim to utilize ensemble learning to make predictions based on three feature levels, we design three sub-models, each intended to process a specific feature. These sub-models incorporate multiple classifiers with different architectures (we also refer to these classifiers as discriminators) to make predictions, which promotes the effectiveness of ensemble learning. The prediction vectors from these sub-models are then merged using a weighted combination approach to train the meta-model, which ultimately determines the final classification of the scene.
 
    To provide interpretability, we developed an extension algorithm that utilizes visual and textual explanations to provide insight into scene recognition predictions. The explanations are composed of key attributes, including the most contributory object categories, frequencies, locations, and textural information. Additionally, we extract scores from sub-models to represent the contribution of these objects towards the prediction, as well as scores that reflect the degree of agreement between each sub-model and the final prediction.
	
    Our proposed method is structured into two sections, the learning algorithm and the explanation generator, which is described hereafter in detail.
	
	\begin{figure*}
		\centering	\includegraphics[width=\textwidth]{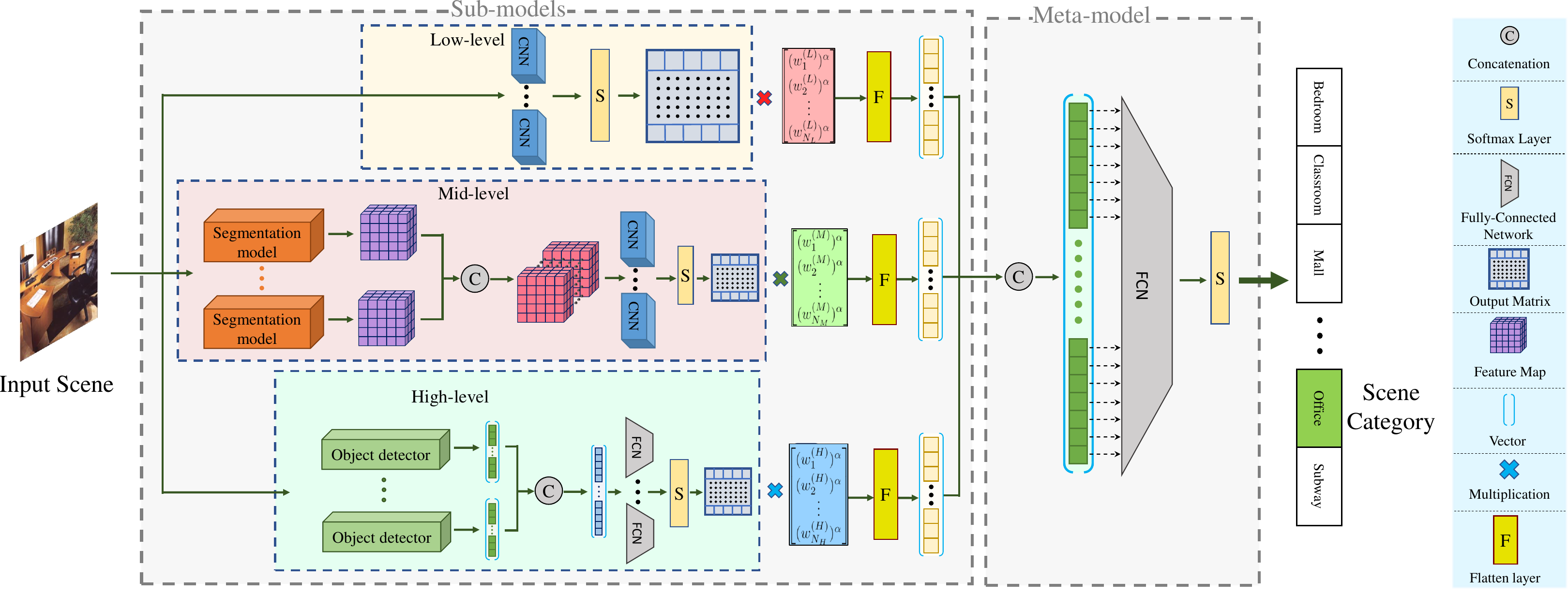}
		\caption{An overview of the proposed framework. The input scene image is first passed into the low-level, mid-level, and high-level sub-models, each producing a prediction matrix. (1) The low-level sub-model uses multiple CNN discriminators to produce the output matrix containing the softmax scores associated with each model; (2) The mid-level model generates multiple segmentation maps via semantic segmentation models, concatenates these maps to build the mid-level representation, and then employs multiple CNN discriminators to generate the prediction matrix; (3) The high-level sub-model constructs the high-level representation by concatenating the output of multiple object detectors and then passes it to the fully-connected discriminators. Finally, the meta-model combines the flattened prediction matrices with a weighted combination scheme, then passes this combination to a fully-connected network to determine the final scene category.}
		\label{framework}
	\end{figure*}

	\subsection{Learning algorithm}
	We describe the learning phase of EnTri in detail in this section, which is illustrated in~\Cref{framework}. As aforementioned, our approach utilizes three sub-models named low-level, mid-level, and high-level, which are designed to process different levels of scene features, ranging from the fundamental pixel level to more abstract and holistic features such as semantic segmentation and object class and frequency, to make a prediction. These predictions are ensembled using a meta-model to recognize the class of the scene. One viable technique to ensure the effectiveness of ensemble learning is to use heterogeneous classifiers that have complementary strengths and weaknesses. To this end, in addition to ensembling branches together, each branch of EnTri also consists of $M$ deep learning models $f(\cdot)$ with different architectures that are used to learn specific patterns in the input scene data and classify them. Given that each branch incorporates multiple discriminator models, their output $F(\cdot)$ can be represented as a matrix $Y \in \mathbb{R}^{N_C \times M}$, where $N_C$ is the number of scene categories, and each entry $y_{i,j}$ represents the probability of the $i^{th}$ model classifying the scene $x$ as belonging to the $j^{th}$ category. We can therefore write:
	
		\begin{equation}
		\label{eq:output}
		F(x)=
		\begin{bmatrix}
			f(x)^{(1)} \\
			f(x)^{(2)} \\
			\vdots \\
			f(x)^{(M)}
		\end{bmatrix}
		= 
		\begin{bmatrix}
			y_{11} & y_{12} & \dots & y_{1N_C} \\
			y_{21} & y_{22} & \dots & y_{2N_C} \\
			\vdots & \vdots & \ddots & \vdots \\
			y_{M1} & y_{M2} & \dots & y_{MN_C}
		\end{bmatrix}
	\end{equation}
 
	\subsubsection{Low-Level sub-model}
	
	As the name suggests, the low-level sub-model branch operates by utilizing perceptual features at the most basic level to differentiate between different scene categories, particularly by focusing on the colors and background of the scene (this can also be referred to as the pixel-level sub-model that operates based on pixel-level features). It incorporates several heterogeneous image classification models $f_L$(low-level discriminators) to make predictions. Each of these models produces a softmax vector $SV = [y_{m,1},y_{m,2}, ...,y_{m,j}]$, where $y_{m,j}$ represents the probability of the scene belonging to the $j^{th}$ category, produced by the $m^{th}$ discriminator model. Therefore, the output of the low-level sub-model $F_{L}(\cdot)$ is represented as~\Cref{eq:output}, where every row represents the softmax output of a discriminator. The pseudocode for the low-level sub-model algorithm is given in~\Cref{alg:img_classification}.
	
	\begin{algorithm}
		\caption{Low-Level sub-model}
		\label{alg:img_classification}
		
		\begin{algorithmic}[1]
			\newcommand{\algorithmicinput}{\textbf{Input:}}
			\newcommand{\Input}{\item[\algorithmicinput]}
			\newcommand{\algorithmicoutput}{\textbf{Output:}}
			\newcommand{\Output}{\item[\algorithmicoutput]}
			
			\Input An input scene $s \in \mathbb{R}^{N_x \times N_y \times 3}$, and $M_L$ different image classification models $f_{L}^{(1)}, f_{L}^{(2)}, \dots, f_{L}^{(M_L)}$
			\Output output probability matrix $Y_L$ over pre-determined categories
			\State $Y_L \gets$ empty matrix of size $M_L \times N_C$
			
			\For{$i \gets 1$ to $M_L$}
			\State Obtain softmax vector of model $i$: $SV_i = f_{L}^{(i)}(x)$ 
			\State Update the $i^{th}$ row of $Y_L$ with $SV_i$
			\EndFor
			\State \Return $Y_L$
		\end{algorithmic}
	\end{algorithm}

	\subsubsection{Mid-Level sub-model}
	The mid-level sub-model serves as a bridge between the low-level sub-model and the high-level sub-model, striking a balance between semantic strength and abstraction for scene recognition (alternatively denoted as the segmentation-level sub-model). In particular, it enables us to capture the shape and category of objects, along with their spatial layout, for prediction. To discriminate between scenes, this sub-model leverages mid-level features, which are constructed through semantic segmentation, a process where the input scene is divided into segments (i.e., the regions covering the objects) based on the meaning of its content. For this purpose, $N_{PM_s}$ segmentation models $PM_{s}$ pre-trained on different datasets are adopted, each of which produces a segmentation map $sm$ with pixel-level labels ($sm = pm_{s}(x) \in \mathbb{R}^{N_x \times N_y \times 3}$). These channels are then concatenated to form the mid-level feature map $MLF \in \mathbb{R}^{N_x \times N_y \times 3N_{PM_s}}$, which captures the spatial properties of objects:
	\begin{equation}
			MLF = \text{concatenate}(sm^{1}, sm^{2}, ..., sm^{N_{PM_{s}}})
	\end{equation}
	 where $sm^{i}$ is the segmentation map produced by the $i^{th}$ semantic segmentation model (The motivation behind using multiple pre-trained segmentation models trained on distinct datasets is to improve scene recognition by covering a wider range of objects. This accounts for situations in which one dataset may not include objects occurring in other datasets). $MLF$ is then fed in parallel into $M_M$ image classification models $F_M$ with different architectures, each trained to generate a softmax probability vector $SV$. Following this, the softmax vectors are merged to generate an output matrix $Y_M$, analogous to the one defined in Equation \ref{eq:output}, which represents the output of the mid-level branch. The algorithm for the mid-level sub-model is outlined in~\Cref{alg:mid-level-submodel}.
	
	\begin{algorithm}
		\caption{Mid-level sub-model}
		\label{alg:mid-level-submodel}
		\begin{algorithmic}[1]
			
			\newcommand{\algorithmicinput}{\textbf{Input:}}
			\newcommand{\Input}{\item[\algorithmicinput]}
			\newcommand{\algorithmicoutput}{\textbf{Output:}}
			\newcommand{\Output}{\item[\algorithmicoutput]}
			
			\Input scene $s \in \mathbb{R}^{N_x \times N_y \times 3}$, $N_{PM_s}$ pre-trained segmentation models $pm_{s}^{1}, pm_{s}^{2}, \dots, pm_{s}^{N_{PM_{s}}}$, and  $M_M$ image classification models $f_{M}^{(1)}, f_{M}^{(2)}, \dots, f_{M}^{(M_M)}$
			\Output Probability vector $Y_M$ over pre-determined categories
            \State $Y_M \gets$ empty matrix of size $M_M \times N_C$
			\State $MLF \gets$ empty matrix of size $N_x \times N_y \times 3N_{PM_{s}}$
			
			\For{$i=1$ to $N_{PM_{s}}$}
				\State $sm_i \gets$ Segmentation map of $s$ using the model $pm_{s}^{i}$
				\State Update $MLF$ with $sm_i$
			\EndFor
			
			\For{$i=1$ to $M_M$}
				\State Obtain softmax vector of model $i$: $SV_i = f_{M}^{(i)}(MLF)$ 
				\State Update the $i^{th}$ row of $Y_M$ with $SV_i$
			\EndFor

			\State \Return $Y_M$
		\end{algorithmic}
	\end{algorithm}

	\subsubsection{High-Level sub-model}
	The high-level sub-model prioritizes semantically strong properties, such as object identity and their frequency of occurrence (alternatively referred to as the object class and frequency-level sub-model). Its primary strength is its ability to maintain robustness and adaptability in the presence of details that might confuse the model. Instead, the model relies on higher-level properties, including the occurrence and count of objects, to represent and classify scenes. For instance, an auditorium and a movie theater may have similar environments that could confuse the classification model, but relying solely on the presence and frequency of different objects without considering their details can aid in enhancing classification accuracy. To identify different scenes, this sub-model utilizes high-level perceptual features, which are extracted and built via $N_{PM_o}$ pre-trained object detection models $PM_o$, each of which is pre-trained on different training datasets to specialize in detecting objects (the reason for using multiple pre-trained object detection models is to enhance scene recognition by incorporating more objects). The output from each object detection model $pm_{o}^{i}$ applied to the scene is a bag-of-objects vector $\mathbf{ov}_i$, defined as a vector where the elements corresponding to the occurring objects equal the count; otherwise, it is zero. This vector is of size $N_{CP_i}$, where $N_{CP}$ represents the number of categories in the dataset for which the $i^{th}$ model has been pre-trained. The vector $ov_i$ for each model $i$ is then concatenated to create a high-level abstract representation $HLF$ of the input scene, which captures the abstract content-level features within the scene:
\begin{equation}
		HLF = \text{concatenate}(ov^{1}, ov^{2}, ..., ov^{N_{PM_o}})
\end{equation}
	Subsequently, the encoded representation $HLF$ is propagated in parallel through $M_H$ fully-connected networks $F_H$ with different architectures to compute the softmax score vectors $SV$. These vectors are then gathered to form the matrix $Y_H$ representing the output of the high-level branch, which mirrors the formulation presented in~\Cref{eq:output}.~\Cref{alg:high-level-submodel} presents the pseudocode for the high-level sub-model algorithm.
	
	\begin{center}
		\begin{algorithm}[H]
			\caption{High-Level sub-model}
			\label{alg:high-level-submodel}
			\begin{algorithmic}[1]
				
				\newcommand{\algorithmicinput}{\textbf{Input:}}
				\newcommand{\Input}{\item[\algorithmicinput]}
				\newcommand{\algorithmicoutput}{\textbf{Output:}}
				\newcommand{\Output}{\item[\algorithmicoutput]}
				
				\Input Input scene $s\in\mathbb{R}^{N_x \times N_y \times 3}$, $N_{PM_o}$ pre-trained object detectors $pm_{o}^{1}, pm_{o}^{2}, \dots, pm_{o}^{N_{PM_o}}$, and $M_H$ fully-connected networks $f_{H}^{(1)}, f_{H}^{(2)}, \dots, f_{H}^{(M_H)}$
				\Output Output probability matrix $Y_H$ over pre-defined categories
                    \State $Y_H \gets$ empty matrix of size $M_H \times N_C$
				\State $HLF \gets$ Empty vector of size $\sum_{i=1}^{M_H} N_{CP_{i}}$
				
				
				\For{$i \gets 1$ to $N_{PM_o}$}
				\State $ov_i \gets$ Bag-of-objects vector of size $N_{CP_{i}}$ for input scene $s$ according to model $pm_{o}^{i}$
				\State Update $HLF$ with $ov_i$
				\EndFor
				
				\For{$i \gets 1$ to $M_H$}
				\State Obtain softmax vector of model $i$: $SV_i = f_{H}^{(i)}(HLF)$ 
				\State Update the $i^{th}$ row of $Y_H$ with $SV_i$
				\EndFor
				
				\State \Return{$Y_H$}
				
			\end{algorithmic}
		\end{algorithm}
	\end{center}

 	\begin{figure*}
		\centering	\includegraphics[width=0.8\textwidth]{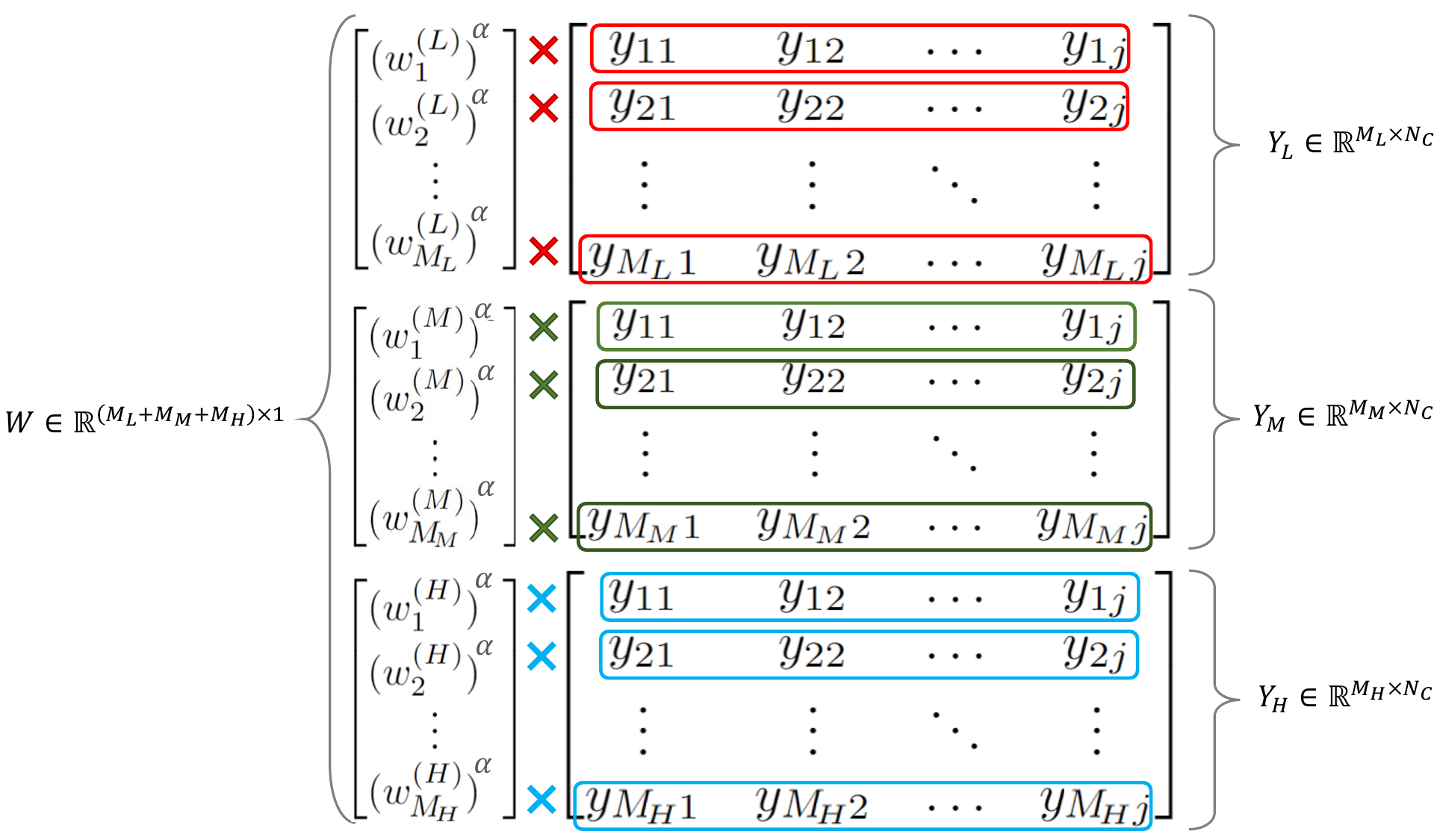}
		\caption{Weighted combination process. The calculated weight values in the constructed weight vectors are multiplied by the entire row from the sub-model matrix that matches the associated discriminator.}
		\label{meta-model}
	\end{figure*}

	\subsubsection{Meta model}
	The meta-model is a fully connected neural network that utilizes output matrices $Y_L \in \mathbb{R}^{M_{L}\times N_C}$, $Y_M \in \mathbb{R}^{M_{M}\times N_C}$, and $Y_H \in \mathbb{R}^{M_{H}\times N_C}$ to learn how to predict based on softmax scores derived from three different levels of perceptual features. Specifically, we devised an approach for combining the predictions as follows. For each sub-model, we construct a weight vector where each weight value is associated with a discriminator in the sub-model. To derive final weight values, we measure the validation accuracy of each discriminator on the validation set, then multiply the results by $100$ (to obtain values greater than $1$) and raise them to the power of an exponent $\alpha$, which ranges from $1$ to $3$. (We did not use trainable weights because the discriminators of every sub-model had training accuracy close to 100\%. When the meta model attempts to train these weights during its training stage, it will be based on the near-perfect training accuracy of the discriminators. This would result in identical weight values, despite the discriminators' different levels of generalization ability and accuracy. To avoid this, weight assignment was based on the accuracy of the discriminators on the validation set). $\alpha$ is a hyperparameter that applies the impact of the discriminator's performance on the weight values to more effectively differentiate between them. The optimal value of $\alpha$ is obtained through the validation stage (The motivation behind using alpha ($\alpha$) is to determine the influence level of each discriminator, ensuring that models with better performance have a proportionally greater impact on the final prediction). We then multiply each calculated weight value with the entire row that aligns with the discriminator's prediction from the sub-model matrix, as illustrated in~\Cref{meta-model} to obtain the weighted prediction matrices $\tilde{Y_L}$, $\tilde{Y_M}$, and $\tilde{Y_H}$. Subsequently, we flatten each matrix into a dimension space of $\mathbb{R}^{1\times MN_C}$. Through the process of flattening, the predictions generated by each model $f(\cdot)$ are arranged in a sequential manner in a vector. These vectors are then concatenated:
	\begin{equation}
		x_{meta} = [~\text{flatten}(\tilde{Y_L}),~\text{flatten}(\tilde{Y_M}),~\text{flatten}(\tilde{Y_H})~]  \in \mathbb{R}^{1\times N_C(M_{L} +M_{M} + M_{H})}
	\end{equation}
	
	where $x_{meta}$ represents the input of the meta-model network. The network then processes this input to produce the final prediction of the input scene category. During the training phase, the meta-model learns to classify scenes based on predictions generated by sub-models and learns optimal network weights for combining these predictions to improve recognition accuracy by combining their complementary strengths.

	\subsection{Explanation generator}
	To embed our framework with interpretability, we devise an algorithm called the Visual Textual Explanation Generator (VTEG) that highlights the scene attributes that contributed to the final prediction of the scene category, including objects, statistics, spatial location, and textural information. VTEG extracts these attributes from the sub-models by considering the prediction score in $SV$ of each sub-model. The extracted attributes, along with the prediction score of each sub-model, are then inserted into predetermined textual formats. The textual explanation indicates why the scene category was decided by providing reasons in three different paragraphs. Each paragraph points out the percentage of agreement between a sub-model and the class generated by the meta-model, which corresponds to the element in the softmax vector generated by the sub-model indicating the probability of input belonging to that class, and also specifies the objects that contributed to the final prediction at their respective feature levels. The process of generating these sentences can be interpreted as a fill-in-the-blank task, where a predetermined sentence with blank spaces is provided and VTEG extracts the attributes to fill in the blanks.

      \begin{figure*}
		\centering	\includegraphics[width=\textwidth]{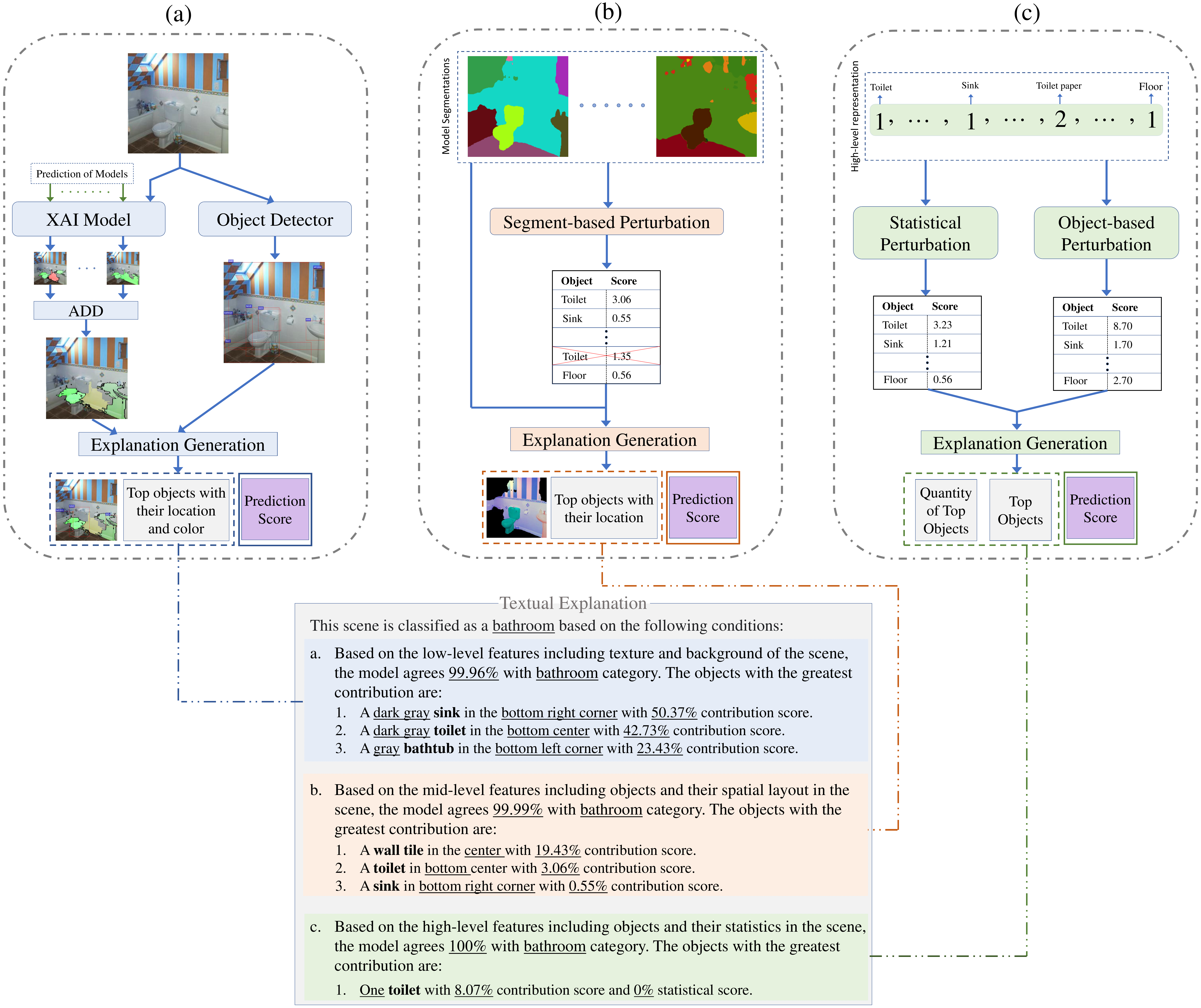}
		\caption{Building blocks of the VTEG algorithm. (a) An explainable AI method generates heatmap explanations for each image classification model, and an object detection algorithm generates an annotated detection image, which is combined with the heatmaps. The top three objects are then extracted using the rate of overlap between the heatmaps and bounding boxes. (b) The segment-based perturbation technique masks different segments of each segmentation map and calculates the importance of objects by evaluating their impact on the prediction score, helping to identify the top three objects. In cases where there are multiple identical objects, the selection process prioritizes the object with the highest score. (c) The statistical and object-based perturbation techniques help identify the top three important objects in the high-level representation and their importance frequency in the scene. Once all the attributes have been extracted, they are inserted into predetermined text formats along with the percentage of agreement between a sub-model and the prediction generated by the meta-model (the high-level section in the textual explanation includes only one object since the other objects scored zero in importance).}
		\label{building}
        \vspace{40pt} 
    \end{figure*}

    \Cref{building} illustrates the building blocks of the explanation generator applied to an MIT67 scene. It can be seen that the textual explanation contains the following contents in each paragraph:

\vspace{10pt}

    \begin{enumerate}[label=\alph*)]
        \item The first paragraph presents the attributes extracted from the low-level sub-model, which first briefly describes the low-level features and then specifies the accuracy of agreement with EnTri's final prediction. Then, indicate the top objects in bulletpoint format with the following content:
        \begin{enumerate}[label=\arabic*.]
                \item The name of the objects
                \item The color of the objects
                \item The object's position in the scene
                \item The contribution score of the object in the prediction
              \end{enumerate}
        \item  The second paragraph focuses on mid-level sub-model attributes, where it overviews the mid-level features, followed by the agreement accuracy with EnTri's prediction:
            \begin{enumerate}[label=\arabic*.]
                \item The name of the objects
                \item The object's spatial position
                \item The contribution score of the object in the prediction
            \end{enumerate}
        \item The third paragraph covers high-level properties, including object categories and statistics, and highlights the agreement's accuracy:
            \begin{enumerate}[label=\arabic*.]
                \item The name of the objects
                \item The frequency of the object's appearance in the scene
                \item The contribution and statistical score of the object in the prediction
            \end{enumerate}
    \end{enumerate}

    \vspace{10pt}
   
    The following sections describe the procedure for extracting the contents listed above from each sub-model.

	\subsubsection{Low-level attribute extractor}
        \label{Low-level extraction}	
	The design of this section is inspired by Aminimehr et al.~\citep{aminimehr4385953tbexplain}. Here, VTEG extracts important low-level attributes in two stages. As depicted in~\Cref{building}(a), in the first stage, an explainable AI method and an object detection algorithm are used to generate heatmap explanations of the outputs and the annotated detection image, respectively. The XAI method is applied to the output of each image classification model $f_{L}$ separately. The resulting heatmaps generated from each model are then added to form a cumulative heatmap. Specifically, given an input image $x$, the explainable AI method generates a heatmap $H(x)$ that represents the contribution of each pixel to the output prediction, while the object detection algorithm generates a set of bounding boxes ${B(x)} $ that represent the locations of objects in the image. In the second stage, the top objects contributing the most to the prediction are extracted according to the cumulative heatmap. Since the cumulative heatmap is the sum of the heatmaps of all of the discriminators, $M_L$ should be integrated into the scoring process to have a score between 0 and 1.

\begin{figure*}
		\centering	\includegraphics[width=0.9\textwidth]{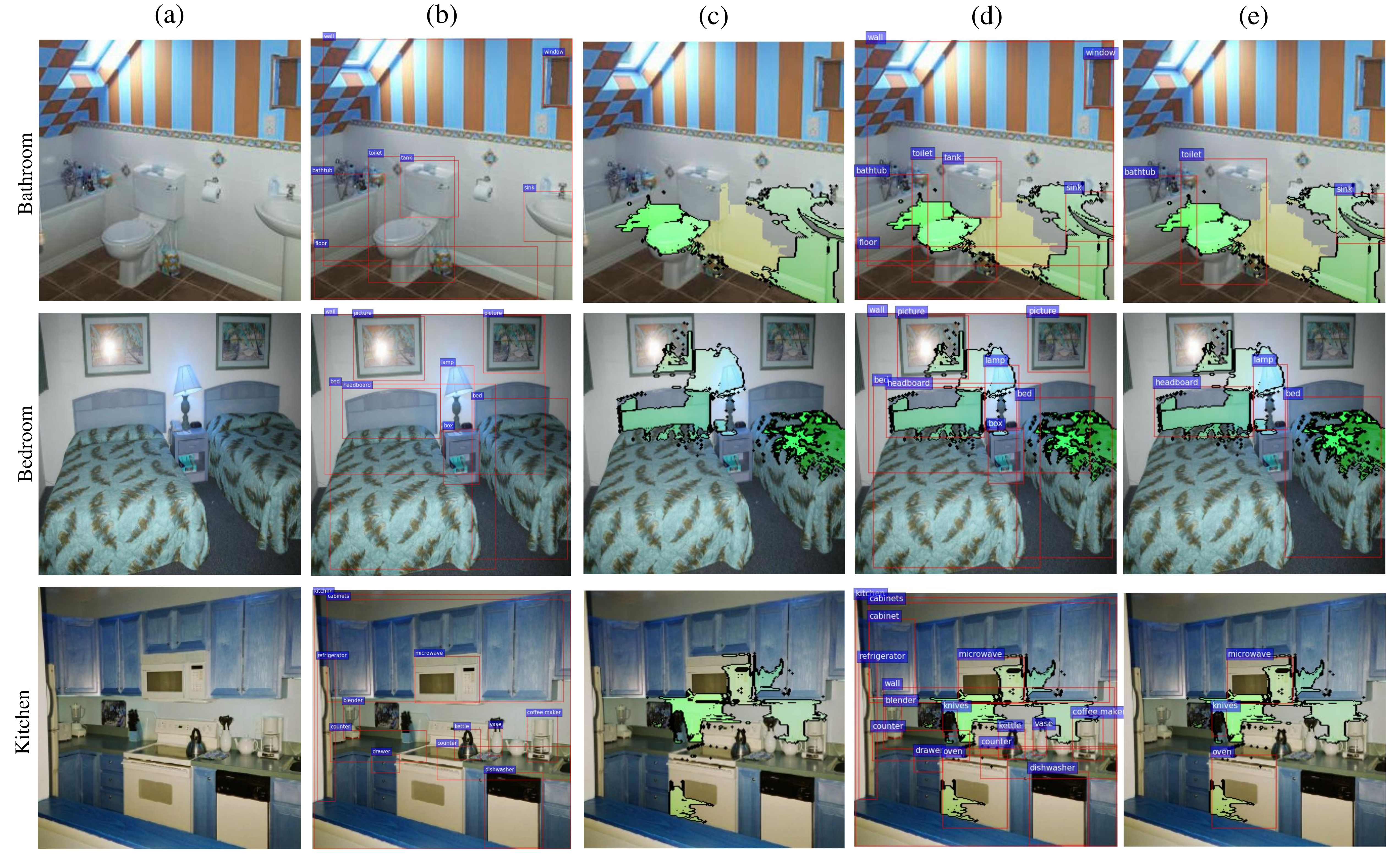}
		\caption{Visualized examples of the extraction process of low-level explanations in VTEG. (a) Input scene. (b) Scene with object bounding boxes. (c) Explainable heatmap (cumulative heatmap). (d) Explainable heatmap with bounding boxes. (e) The heatmap with bounding boxes of the top three objects with the greatest contribution.}
		\label{low-explanation}
    \end{figure*}

    To extract important objects, we divide the sum of the cumulative heatmap pixel values in each bounding box by the area of the bounding box multiplied by the number of discriminators $M_L$ (Since the cumulative heatmap is the summation of the heatmaps generated by the discriminators, $M_L$ should be integrated into the scoring process to produce a value ranging between $0$ and $1$). To do this, each bounding box $b_i(x)$ is assigned a score $score(b_i(x))$:
	\begin{equation}
		score(b_i(x)) = \frac{heat(b_i(x))}{area(b_i(x)) M_{L} }
	\end{equation}
	where $heat(b_i(x))$ is the sum of the cumulative heatmap pixel values within the bounding box $b_i(x)$, and $area(b_i(x))$ is the area of the bounding box. The three objects with the highest scores are considered to have made the greatest contribution. The visual examples presented in~\Cref{low-explanation} demonstrate VTEG's low-level attribute extraction on samples from the MIT67 dataset.

    To identify the color of the important object, our solution is to consider the color of the pixel located at the center of the object's bounding box as the color of the object. 

    In order to detect the position of an object, the image undergoes a partitioning procedure, where it is divided into $9$ distinct sections. These sections can be regarded as a $3\times3$ grid with $9$ cells. The object's position is then determined by identifying the section in which the centroid of its bounding box is located.

	\subsubsection{Mid-level attribute extractor}
	\label{Mid-level extraction}
	As illustrated in~\Cref{building}(b), given a set of segmentation maps $SM$, a segment-based perturbation technique is employed that involves masking different segments (i.e., removing the objects) of each segmentation map $sm_{i}\in SM$ and assessing the importance of objects, which is inspired by the work of Anjomshoae et al.~\citep{anjomshoae2021context}. In particular, VTEG applies a mask to each segment of the segmentation maps separately by setting its pixels to the background segment values, and then passes its concatenation with other segmentation maps to the discriminator models to determine how much the masked objects impact the mid-level sub-model's prediction score. After measuring the prediction with each object masked, the impact of each object on the model's prediction is calculated using the following equation:
	\begin{equation}
		score_{j} = \frac{\max_{i=1}^{\lvert Q \rvert} (F_{M}(MLF_{Q-i})) - F_{M}(MLF_{Q-{j}}) }{\max_{i=1}^{\lvert Q \rvert} (F_{M}(MLF_{Q-i})) - \min_{i=1}^{\lvert Q \rvert} (F_{M}(MLF_{Q-i}))}
	\end{equation}

	\begin{figure*}
		\centering	\includegraphics[width=0.8\textwidth]{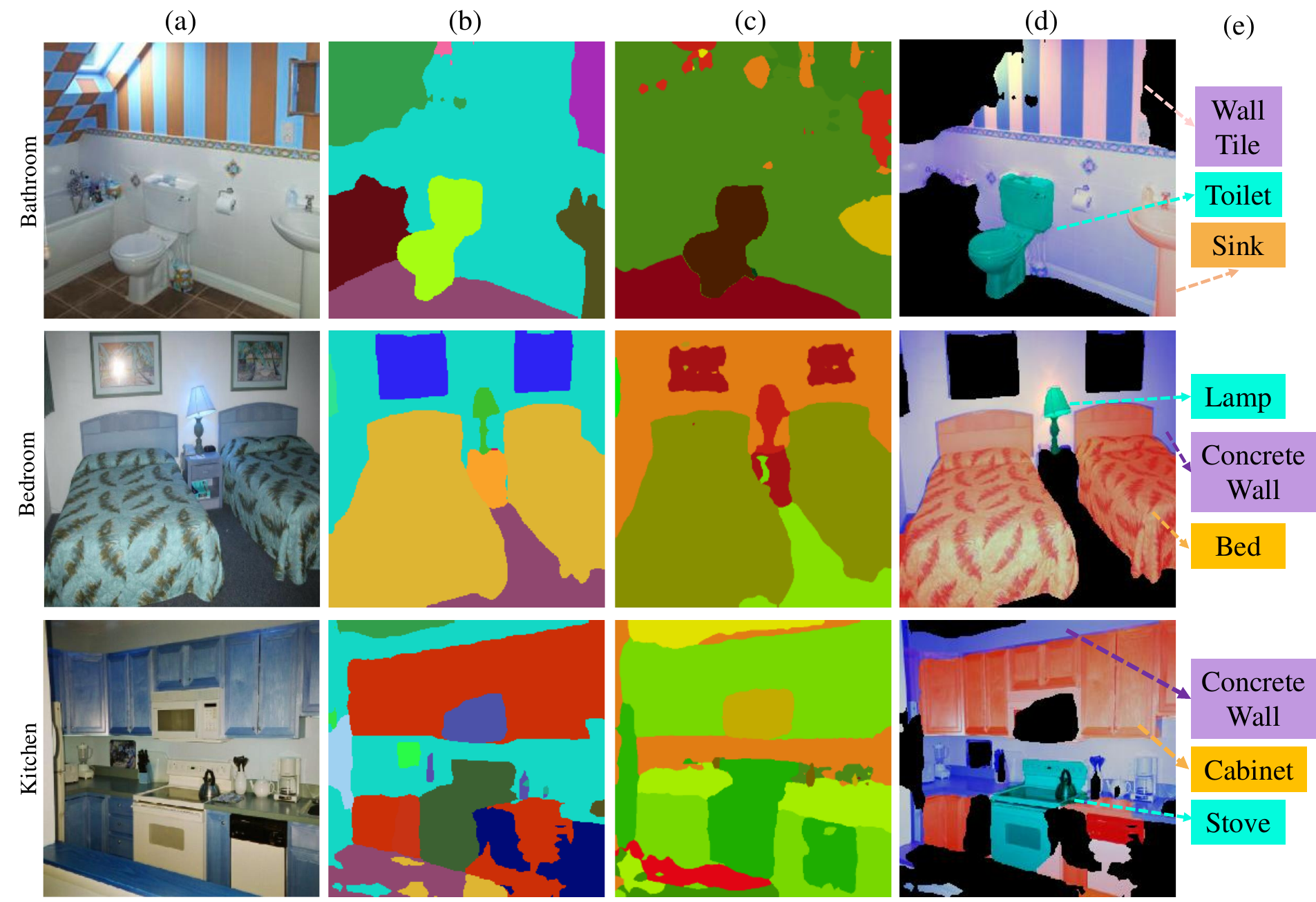}
		\caption{Visualized examples of extracting mid-level explanations with VTEG. (a) Input scene. (b) Segmentation map of the first model. (c) Segmentation map of the second model. (d) Masked segmentation map highlighting the important objects. (e) Important objects (attributes) extracted based on mid-level information.}
		\label{mid-explanation}
	\end{figure*}

	where $score_{j}$ denotes the impact of object $j$ on the prediction score, $Q$ is the list of all objects in the $SM$, and $MLF_{Q-i}$ represents the mid-level features witholding the object $i$ from $Q$ (setting the pixels of object $i$ in the $SM$ to the background segment value). The top three objects with the highest $score_{j}$ are considered the most important attributes. In the case of multiple identical objects over different segmentation maps, our selection process prioritizes the object with the higher score (e.g., a toilet object is present in two segmentation maps, so the one with a lower score will be removed).~\Cref{mid-explanation} illustrates some visualized sample results of VTEG's mid-level attribute extraction, including segmentation maps and the top three most contributing objects, on the MIT67 dataset. To detect the spatial location of the top objects, the same method as described in~\Cref{Low-level extraction} is employed.
	
	\subsubsection{High-level attribute extractor}
	To identify the key components in the high-level sub-model's prediction process, we utilize a technique called object-based perturbation similar to the method described in~\Cref{Mid-level extraction}, which is shown in~\Cref{building}(c). This perturbation strategy involves zeroing out the values corresponding to the scene objects that comprise the high-level representation $HLF$ and subsequently measuring their impact on the prediction score of the sub-model. Specifically, for an object $i$, we obtain its importance score by computing:
	
	\begin{equation}
          Contribution~Score(i) = \frac{\max_{i=1}^{\lvert O \rvert} (F_{H}(HLF_{O-i})) - F_{H}(HLF_{O-{j}})}{\max_{i=1}^{\lvert O \rvert} (F_{H}(HLF_{O-i})) - \min_{i=1}^{\lvert O \rvert} (F_{H}(HLF_{O-i}))}
	\end{equation}

	where $HLF_{O-i}$ represents the high-level features with the elements corresponding to object $i$ zeroed out, and $O$ is the list of all objects in the $OV$. The set of top three objects is subsequently chosen for inclusion in the textual explanation. Each object within this set is unique and does not repeat. If any identical objects are present in this top three set, we choose the one with the highest score and attempt to substitute the remaining ones with different, non-identical objects with lower scores.
 
    In addition to the other metric, we also introduce a statistical score that represents the impact of the frequency of a given object $i$ on the performance, independent of its presence in the scene. As we solely use the top three objects with the highest contribution score for generating explanations, we restrict our statistical score calculations to these specific objects. To assess the statistical importance of objects, we devise a statistical perturbation-based technique as follows. For each object $i$, we calculate the score for each particular quantity of that object with the following equation (zero is also considered):
 
	\begin{equation}
          QS_{i}(c_k) = \frac{\max_{k=0}^{m_i} (F_{H}(HLF_{k})) - (F_{H}(HLF_{c_k}))}{\max_{k=0}^{m_i} (F_{H}(HLF_{k})) - \min_{k=0}^{m_i} (F_{H}(HLF_{k}))}
	\end{equation}

        where $QS_{i}(c_k)$ is the quantity score for object $i$ with the quantity of $c_k$ instances in the representation, and $m_i$ is the maximum count of object $i$ in the representation (i.e., the frequency of the object in the original $HLF$ representation). To measure the statistical score for the object $i$, we start with its maximum quantity $m_i$, then gradually decrease its count by $1$ and measure the difference of successive quantity scores until we reach the minimum count $0$ for that particular object. We then calculate the mean of all the subtractions of successive quantity scores to obtain the statistical importance score of the associated object:

	\begin{equation}
		Statistical~Score(i) = \frac{1}{m_{i-1}} \sum_{k=1}^{m_{i-1}} ~ \lvert ~ QS_{i}(m_i - k) - QS_{i}(m_i-k-1)) ~ \rvert
	\end{equation}

	\section{Experiments}
	In this section, we discuss the experiments on the proposed method carried out using the two benchmark datasets, namely MIT67 and SUN397. We compare the proposed method with traditional and modern models under the same class supervision.

	\subsection{Benchmark datasets}

	\subsubsection{MIT67}
	MIT67 dataset~\citep{quattoni2009recognizing} contains 15620 scene images grouped into 67 different categories, each representing a subset of five environments: public places, working places, home locations, stores, and leisure sites. The images are all in JPEG format, have a resolution of 256x256 pixels, and were collected from online image search engines and photo sharing websites. We followed the standard data split, allocating 80 images for training and 20 for testing from each category. We further partitioned the training set in an 80:20 ratio for training and validation, respectively. Several scene images belonging to the dataset are illustrated in~\Cref{dataset}.
	
	
	\subsubsection{SUN397}
	The SUN397 dataset~\citep{xiao2010sun} is a comprehensive collection of 108,754 images depicting 397 distinct scene categories. Each category features at least 100 images, although the number of images may vary across different categories. It is widely used in the field of computer vision for a variety of applications, such as scene recognition, image classification, and object detection. Due to the large number of images and scene categories, it provides a challenging benchmark for researchers to test and evaluate their algorithms. The dataset includes an evaluation protocol that outlines how to divide the data into separate training and validation sets. We used the standard data split, assigning 50 images per category for training and 50 for testing, and divided the training set into an 80:20 ratio for training and validation.~\Cref{dataset} presents several examples of scene images from the dataset.
	
	\subsubsection{UIUC8}
\begin{figure*}[!t]
		\centering	\includegraphics[width=0.8\textwidth]{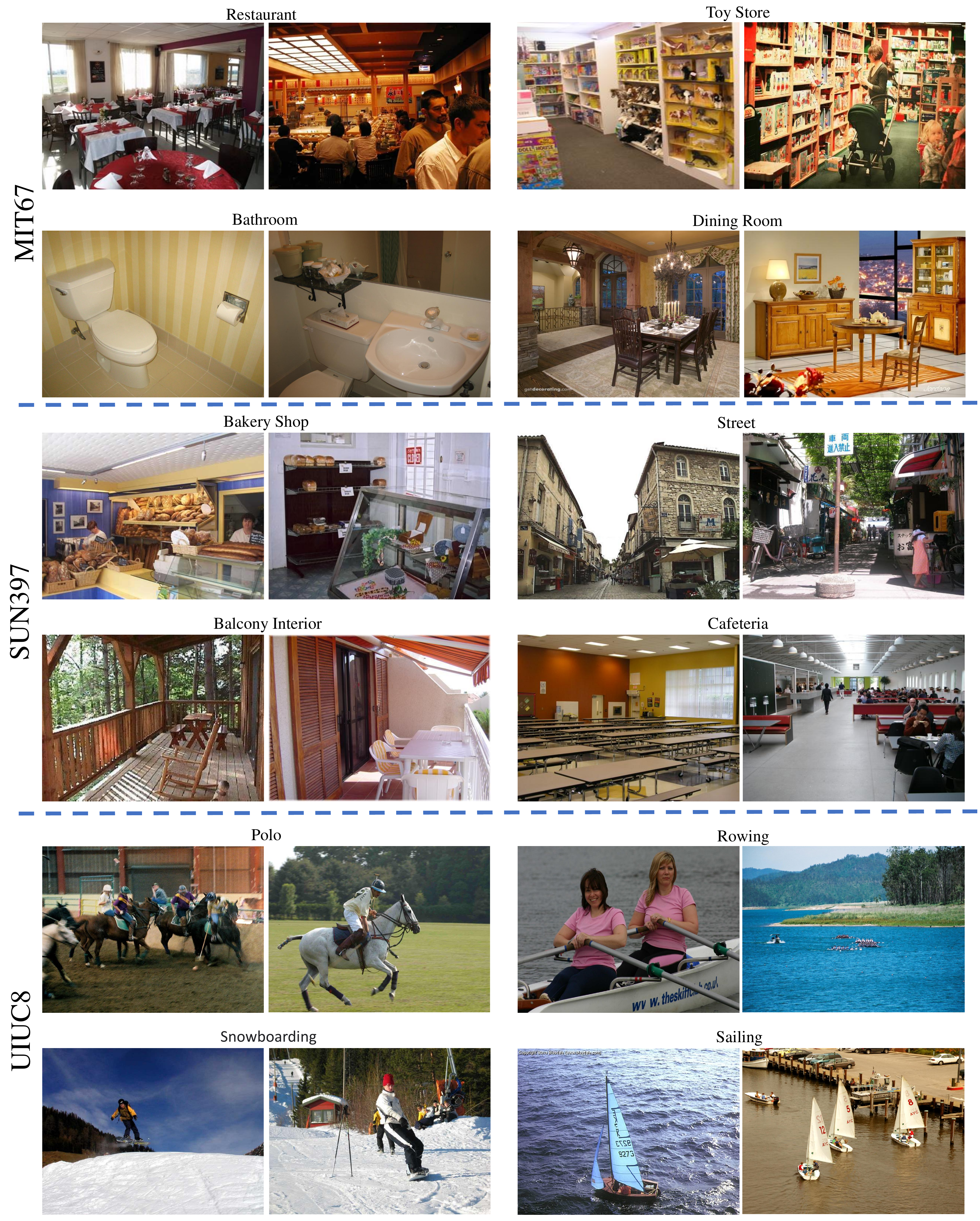}
		\caption{A selection of images from MIT67~\citep{quattoni2009recognizing}, SUN397~\citep{xiao2010sun}, and UIUC8~\citep{li2007and} datasets, providing a glimpse into the nature of the data included in each dataset.}
		\label{dataset}
	\end{figure*}

	The UIUC8 event dataset~\citep{li2007and} contains 1579 images of 8 different sports categories, including bocce, croquet, polo, rowing, snowboarding, badminton, sailing, and rock climbing. Each event category contains between 137 and 250 images collected from the internet. We followed the standard split, which includes 70 training and 60 test samples. We randomly selected the data for each set since no protocol was provided for the selection process. We also divided the training set into an 80:20 ratio for training and validation.~\Cref{dataset} provides some examples of the dataset.


	\subsection{Implementation protocol}
        \label{Implementation}
        
	\textbf{Network}: We used two discriminator models for each sub-model ($M_L=2$, $M_M=2$, and $M_H=2$). For the low-level submodel, we chose to fine-tune ConvNeXt-B~\cite{liu2022convnet} and EfficientNetV2-L~\cite{tan2021efficientnetv2} with pre-trained weights on the Imagenet dataset~\cite{deng2009imagenet}. The mid-level sub-model uses two semantic segmentation networks ($N_{PM_s} = 2$), DeepLab v2~\cite{chen2017deeplab} with a ResNet-101 backbone pre-trained on the COCO-Stuff dataset~\cite{caesar2018coco} and PSPNet 50~\cite{zhao2017pyramid} pre-trained on the ADE20k dataset~\cite{zhou2019semantic} and trained a DenseNet-201~\cite{huang2017densely} and Inception-ResNet-v2~\cite{szegedy2017inception} as the discriminators. Finally, the high-level sub-model is implemented with Faster R-CNN~\cite{ren2015faster} pre-trained on the Visual Genome dataset~\cite{krishna2017visual}, Mask R-CNN X-101-32x8d FPN~\cite{he2017mask} pre-trained on the LVIS v0.5 dataset~\cite{gupta2019lvis}, Yolov5m~\cite{glenn_jocher_2022_7347926} pretrained on the Objects365 dataset~\cite{shao2019objects365}, YOLOv5l pretrained on the Open Images V6 dataset~\cite{kuznetsova2020open}, and Yolov5x6 pretrained on the COCO dataset~\cite{lin2014microsoft} as the object detectors, and has two fully-connected networks with one hidden layer with sizes of 4096 and 8192, respectively, as descriminators.
 
	\textbf{Training}: The computational resources for this study are 12 GB of system RAM and a Nvidia Tesla T4 GPU with 16 GB of memory (GDDR6).
 
	\begingroup
	\def\arraystretch{1.2}%
	
	\subsection{Ablation study}
	
	The aim of this section is to assess the influence of individual branches on the performance of our proposed method. We evaluate the influence of each branch architecture, including low-level, mid-level, and high-level discriminators. All the ablation studies are carried out utilizing the MIT67 dataset, which is summarized in~\Cref{tab:ablation}. The table shows the performance metrics for different cases. The comparison is as follows:

\vspace{10pt}
	\begin{enumerate}
		\item \textbf{Cases 1-3}: In these cases, only one of the sub-models is utilized for prediction. Case 1, which solely uses low-level features, demonstrates the highest performance, followed by Case 3, which only incorporates the high-level sub-model. On the other hand, Case 2, which considers mid-level features, achieves the lowest performance among all cases. 
		\item \textbf{Cases 4-6}: In these cases, EnTri incorporates a combination of two levels of sub-models. The highest performance is achieved by Case 5, which utilizes both low-level and high-level sub-models. Following closely is Case 4, which takes into account both low-level and mid-level sub-models. On the other hand, case 6, which uses mid-level and high-level features, demonstrates the weakest performance among the cases. This suggests that low-level and high-level sub-models have a higher impact compared to the mid-level sub-model.
            \item \textbf{Case 7}: All three levels of sub-models (low, mid, and high) are incorporated in this case, resulting in the highest performance and demonstrating their complementary roles for the recognition task.
	\end{enumerate}
	
	\begingroup
	\def\arraystretch{1.2}%
	
	\begin{table}[h]
		\centering
		\caption{Ablation studies on the MIT67 dataset demonstrating the effect of different sub-models evaluated in terms of Top@1, Top@2, and Top@5 accuracy.}
		\label{tab:ablation}
		\begin{tabular}{V{4}c|c|c|c|c|c|cV{4}}
			\hlineB{4}
			\textbf{Case} & \textbf{Low-level} & \textbf{Mid-level} & \textbf{High-level} & \textbf{Top@1 (\%)} & \textbf{Top@2 (\%)} & \textbf{Top@5 (\%)}\\
			\hlineB{4}
			1 & $\checkmark$ & $\times$ & $\times$ & 84.47 & 90.75 & 94.18 \\
			\hline
			2 & $\times$ & $\checkmark$ & $\times$ & 66.49 & 76.57 & 85.45 \\
			\hline
			3 & $\times$ & $\times$ & $\checkmark$ & 77.31 & 86.04 & 92.61 \\
			\hline
			4 & $\checkmark$ & $\checkmark$ & $\times$ & 85.52 & 92.24 & 95.52  \\
			\hline
			5 & $\checkmark$ & $\times$ & $\checkmark$ & 86.19 & 92.84 & 95.97 \\
			\hline
			6 & $\times$ & $\checkmark$ & $\checkmark$ & 79.62 & 87.76 & 93.06 \\
			\hline
			\textbf{7} & \textbf{$\checkmark$} & \textbf{$\checkmark$} & \textbf{$\checkmark$} & \textbf{87.69} & \textbf{93.58} & \textbf{97.54} \\
			\hlineB{4}
		\end{tabular}
	\end{table}
	\endgroup

	\begin{figure*}[!t]
		\centering	\includegraphics[width=\textwidth]{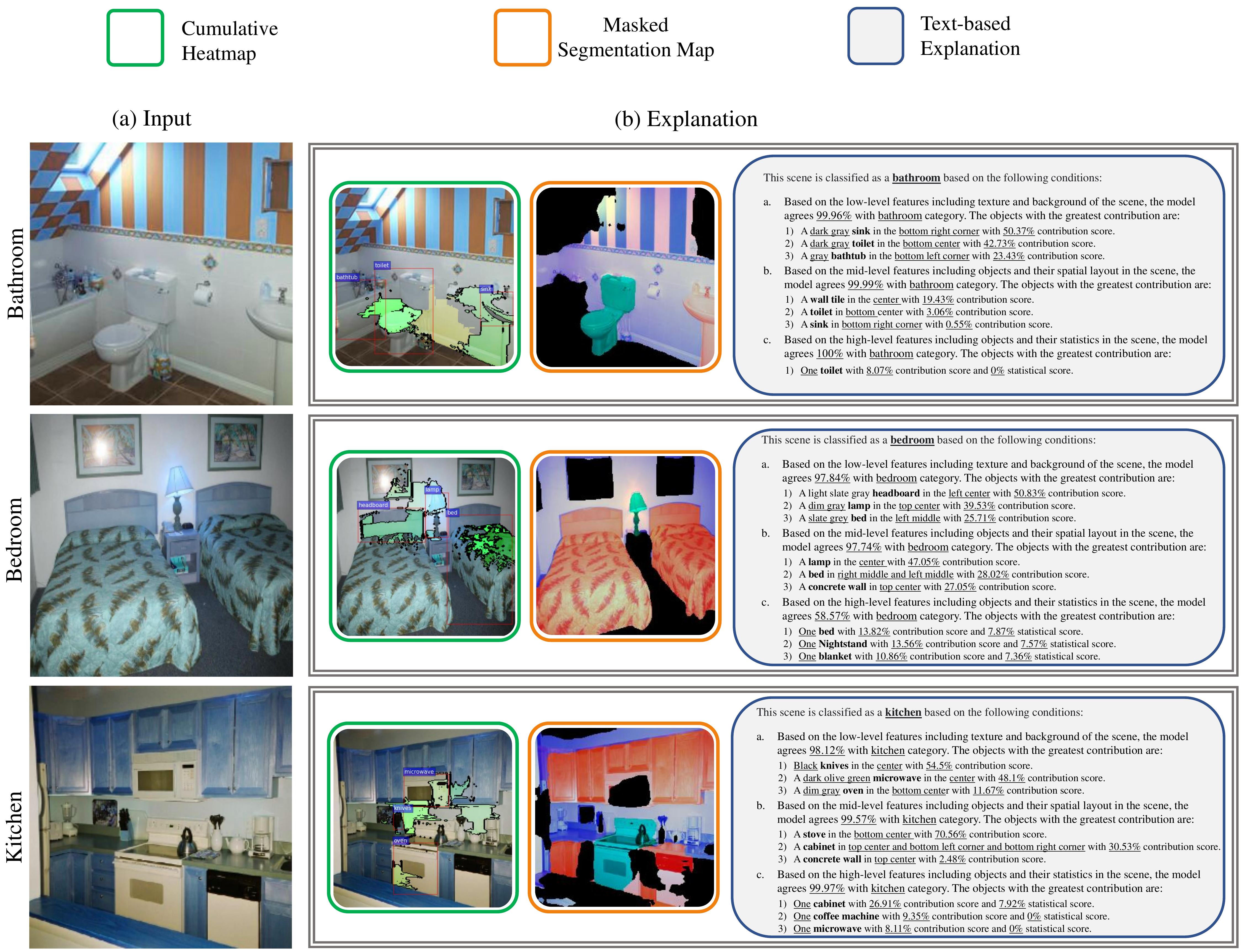}
		\caption{Visual and textual explanations generated by VTEG on MIT67. (a) Input scene. (b) Visual explanations highlight influential attributes and objects in the recognition model's prediction process based on the low-level, mid-level, and high-level features, with the cumulative heatmap indicating color and textural information, the masked segmentation map indicating spatial layout, and the textual explanation showcasing the prediction scores of each sub-model and important objects behind their reasoning.}
		\label{explanation-results}
	\end{figure*}

    Moreover, these textual explanations present the degree of agreement between the predictions made using the three features through ensemble learning and the final prediction of our framework.
	
	\subsection{Explanation results and analysis}
	
	To visualize the quality and intelligibility of the explanations produced by the proposed VTEG, we provided illustrative examples of the visual and textual explanations on different MIT67 scenes in~\Cref{explanation-results}. It can be seen from the examples that the visual explanations attempt to highlight the most influential attributes of the scene in the prediction process of the recognition model, with the cumulative heatmap highlighting the regions that contributed to the prediction based on low-level feature and the masked segmentation map highlighting the important objects that contributed based on mid-level feature. These visual explanations are complemented by the textual explanation, which not only highlights crucial attributes such as object categories, frequencies, locations, and textural cues but also offers scores that represent the contribution of these objects towards the prediction. This textual explanation increases the certainty and confidence of users in their interpretation of the reasoning behind the prediction. Moreover, the ability to align the textual information with the visual contents allows for a clearer comprehension of the process and facilitates the diagnosis of the system at the modular level. By analyzing the scene and the confidence score of each sub-model, users can gain a deeper understanding of which type of information in the scene has the greatest impact on recognition performance. Such insights can help users identify areas for modification and adjustment and provide guidance in answering questions such as:

 \vspace{10pt}

	\begin{enumerate}
		\item \textbf{Data}: Does the data need denoising and quality improvements? Which level of feature is better for differentiating between two similar categories? Which objects in different categories have a significant influence on the model's ability to distinguish between different scene categories? In what kinds of scenes does the quantity of objects highly affect the model's recognition performance? 
		\item \textbf{Network}: Why did the model predict that category? Is the model's final prediction reliable? How do sub-models differentiate when recognizing an object as important content in a scene? Which sub-model is more reliable in a particular category? At what level of representation does the system require improvement? What factors contribute to the generalizability and robustness of these systems, and how can we optimize these factors to improve the model's performance?
	\end{enumerate}

 \vspace{10pt}
	
	Overall, these explanations serve as valuable guidance for users seeking solutions to these questions while also enhancing the transparency and trustworthiness of recognition systems.

	\subsection{Comparison with state-of-the-art methods}
	
	\begin{table}[h]
		\centering
		\caption{Statistical comparison with different state-of-the-art methods on the MIT67 dataset~\citep{quattoni2009recognizing}. The bold numbers represent the best results compared to other methods in this table.}
		\label{tab:results_mit67}
		
			\begin{tabular}{|l|c|c|}
				\Xhline{3\arrayrulewidth}
				\rowcolor[RGB]{255,222,153}
				\multicolumn{3}{|c|}{\textbf{Traditional Approaches}} \\ \Xhline{3\arrayrulewidth}
				\textbf{Method} & \textbf{Year} & \textbf{Accuracy (\%)} \\ 
				\hline
				CENTRIST~\cite{wu2010centrist} & 2010 & 36.9  \\
				Object Bank~\cite{li2010object} & 2010 & 37.6  \\       
                IFV~\citep{juneja2013blocks} & 2013 & 60.77 \\
                ISPR~\citep{lin2014learning} & 2014 & 50.10 \\
                ISPR + IFV~\citep{lin2014learning} & 2014 & 68.50 \\
				\Xhline{3\arrayrulewidth}
				\rowcolor[RGB]{200,255,255}
				\multicolumn{3}{|c|}{\textbf{Modern Approaches}} \\ \Xhline{3\arrayrulewidth}
				\textbf{Method} & \textbf{Year} & \textbf{Accuracy (\%)} \\ \hline
                
				MOP-CNN~\cite{gong2014multi} & 2014 & 68.9 \\
				DAG-CNN~\cite{yang2015multi} & 2015 & 77.5 \\
                 DRCF~\citep{khan2016discriminative} & 2016 & 71.80 \\
				Multi-scale CNNs~\cite{herranz2016scene} & 2016 & 86.04 \\
				VSAD~\cite{wang2017weakly} & 2017 & 86.2 \\
				Soft Combination~\citep{bai2018softly} & 2018 & 82.31 \\ 
				SDO~\cite{cheng2018scene} & 2018 & 86.76 \\
                
				Semantic-Aware~\cite{lopez2020semantic} & 2020 & 87.10 \\ 
                    EML-ELM~\citep{wang2022embedding} & 2022 & 87.10 \\ \hline
				\textbf{EnTri} & \textbf{2023} & \textbf{87.69}  \\ \hline
			\end{tabular}
	
	\end{table}

	\begin{table}[h]
	\centering
	\caption{Statistical comparison with different state-of-the-art methods on the SUN397 dataset~\citep{xiao2010sun}.}
	\label{tab:results_sun397}		

			\begin{tabular}{|l|c|c|}
                    \Xhline{3\arrayrulewidth}
				\textbf{Method} & \textbf{Year} & \textbf{Accuracy (\%)} \\ \hline
				MOP-CNN~\cite{gong2014multi} & 2014 & 51.98 \\
				DAG-CNN~\cite{yang2015multi} & 2015 & 56.2 \\				
				Multi-scale CNNs~\cite{herranz2016scene} & 2016 & 70.17\\
				VSAD~\cite{wang2017weakly} & 2017 & 73.00 \\
				Soft Combination~\citep{bai2018softly} & 2018 & 63.24 \\ 
				SDO~\cite{cheng2018scene} & 2018 & 73.41\\
				Semantic-Aware~\cite{lopez2020semantic} & 2020 & 74.04 \\ 
                AdaNFF~\citep{zou2022adanff} & 2022 & 74.18 \\ \hline
    	
				\textbf{EnTri} & \textbf{2023} & \textbf{75.56} \\ \hline
			\end{tabular}
	\end{table}	
	
	\begin{table}[h]
	\centering
	\caption{Statistical comparison with different state-of-the-art methods on the UIUC8 dataset~\citep{li2007and}.}
	\label{tab:results_uiuc8}	

			\begin{tabular}{|l|c|c|}
				\hline
				\textbf{Method} & \textbf{Year} & \textbf{Accuracy (\%)} \\ \hline
                ISPR+IFV~\citep{lin2014learning} & 2014 & 92.08 \\
                DRCF~\citep{khan2016discriminative} & 2016 & 98.70 \\
			RLML-LSR~\citep{wang2021robust} & 2021 & 98.33 \\ 
                EML-ELM~\citep{wang2022embedding} & 2022 & 98.69 \\ 
                AdaNFF~\citep{zou2022adanff} & 2022 & 99.17 \\ \hline
				\textbf{EnTri} & \textbf{2023} & \textbf{99.17} \\ \hline
			\end{tabular}
		
	\end{table}

	The recognition results are summarized in~\Cref{tab:results_mit67,tab:results_sun397,tab:results_uiuc8}. The Top@1 accuracy for all previous methods listed in the table has been taken from their associated papers. The methods based on traditional hand-engineered and modern deep-learned features are presented in separate sections. It is apparent that deep learning-based methods are superior compared to methods based on hand-engineered features since CNN architectures are capable of handling the complexity and ambiguity of scene data efficiently. As mentioned in the~\Cref{modern approaches}, our model belongs to the modern category as it uses CNNs and fully connected networks to carry out feature extraction and training. From the tables, it is evident that our approach enhances accuracy by mitigating intra-class variations and inter-class similarities using three levels of perceptual features. Our best model is based on the implementation scheme described in~\Cref{Implementation}. In comparison to the related work on MIT67~\citep{quattoni2009recognizing}, EnTri surpassed SemanticAware~\cite{lopez2020semantic} and ELM-ELM~\citep{wang2022embedding} by $0.59$ accuracy.

    While EnTri is more focused on visual data prone to inter-class similarity and intra-class variations, it also shows equivalent performance to state-of-the-art methods in terms of recognition accuracy on the UIUC8 dataset~\citep{li2007and}, which is comparably more simple. As shown in~\Cref{tab:results_uiuc8}, EnTri achieves an accuracy of $99.17$ equal to the AdaNFF model~\citep{zou2022adanff}.
    
    To make more convincing results, we carried out comparative experiments on the SUN397 dataset~\citep{xiao2010sun} summarized in~\Cref{tab:results_sun397}, which is considered to be more challenging than the MIT67. Its larger scale makes training and evaluation of deep-scene models more practical while accentuating the challenges of inter-similarity and intra-variation. EnTri outperformed AdaNFF by $1.38$ accuracy on this dataset and yielded greater improvements in recognition accuracy on SUN397 than on the MIT67 and UIUC8 datasets, which implies that EnTri alleviates the challenges posed by intra-class variations and inter-class similarity.

	\begin{figure*}[!t]
		\centering	\includegraphics[width=0.8\textwidth]{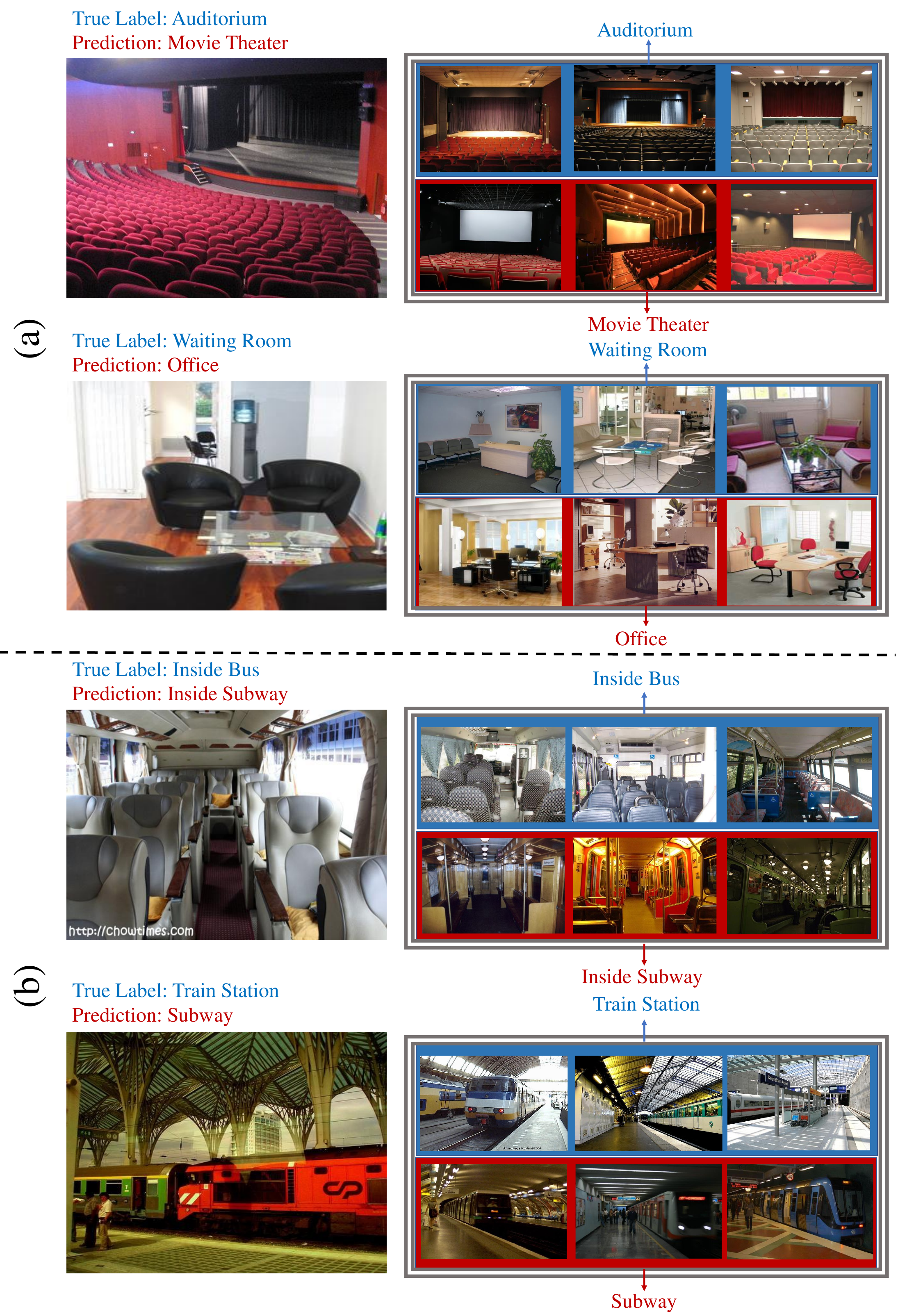}
		\caption{The visualized examples of misclassifications by either (a) the low-level sub-model or (b) the object-level sub-models (mid and high-level). The left column represents the scene being misclassified, and the right column includes scenes from both the misclassified category and the true label category for comparison purposes.}
		\label{discussion}
	\end{figure*}

 	\begin{figure*}
		\centering	\includegraphics[width=\textwidth]{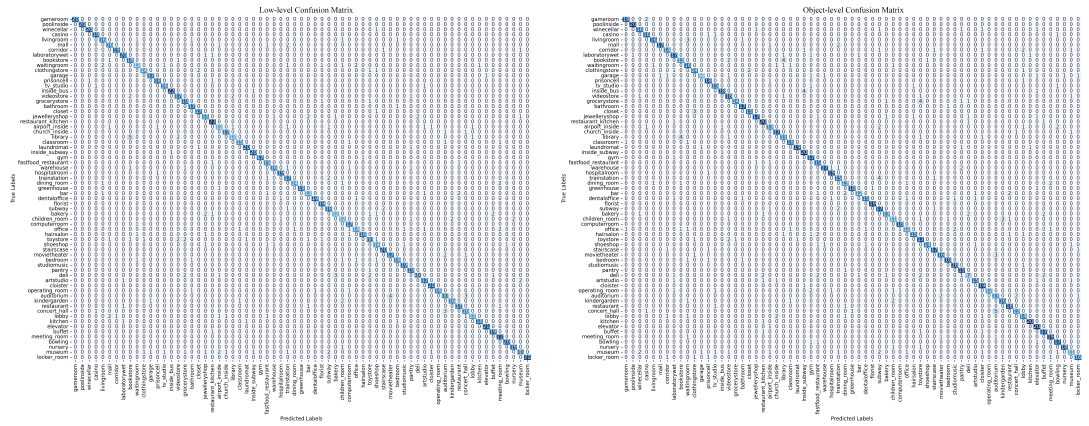}
		\caption{Confusion matrices for evaluating the low-level and object-level sub-models on the MIT67.}
		\label{LO-confusion}
	\end{figure*}
 
 	\begin{figure*}
		\centering	\includegraphics[width=\textwidth]{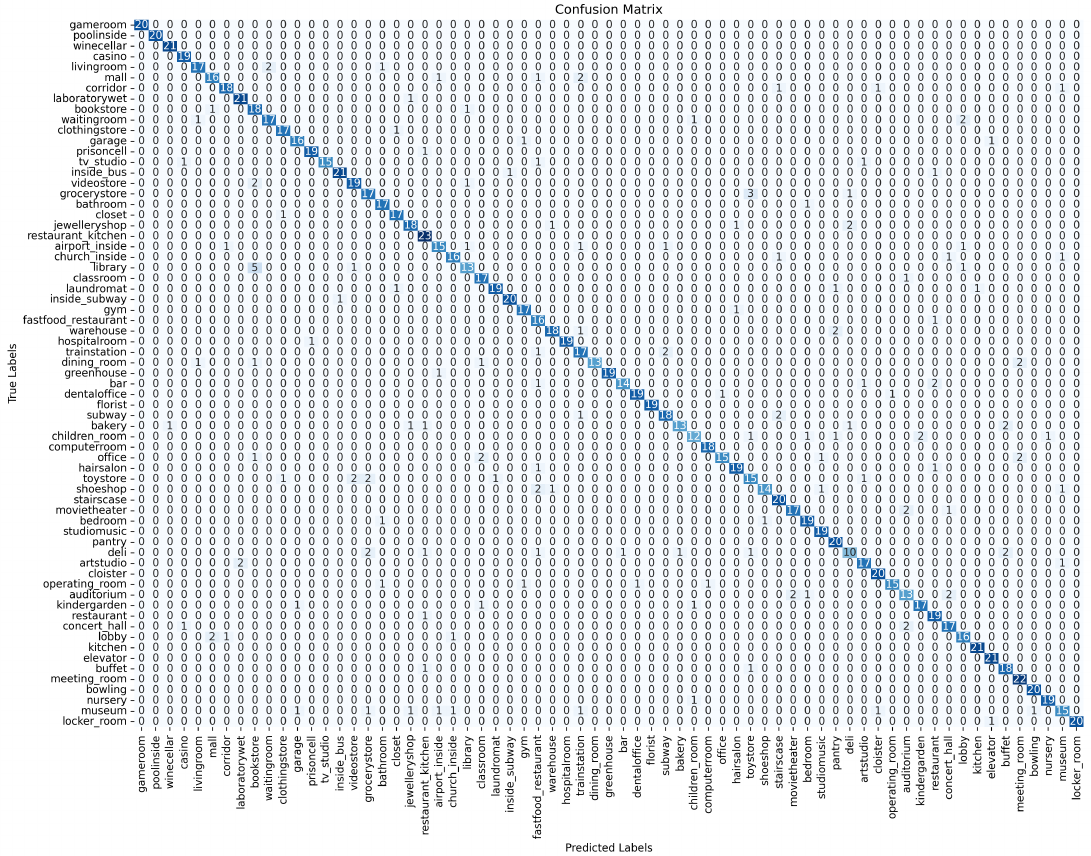}
		\caption{Confusion matrix of the evaluation on the MIT67 dataset using EnTri.}
		\label{confusion}
	\end{figure*}

	\subsection{Discussion}
         In this section, we present additional analysis and experiments to further validate the efficiency of our proposed methods. Our focus is on the influence of different levels of sub-model branches on recognition performance, based on experiments conducted on the MIT67 dataset. To help us describe the effectiveness of each sub-model and provide insights into the strengths and weaknesses of the model's accuracy in classifying individual categories, we have included the confusion matrices of evaluating the low-level and object-level sub-models (mid and high-level combined) and EnTri (including all of the sub-models) on the MIT67 dataset in~\Cref{LO-confusion,confusion}. Additionally, we also provide visual examples in~\Cref{discussion} to demonstrate the effectiveness of our sub-models.
         
        In each confusion matrix, a notable proportion of the classes are accurately predicted with high precision and recall. However, it is apparent that the low-level sub-model falls short compared to object-level sub-models and EnTri when it comes to predicting images of different categories that have similar low-level details but varying object categories and statistics. These details are the scene's environment, background, color, and texture. For example, for the "auditorium" category, EnTri (all of the sub-models) accurately predicts two more instances ($\text{True Possitive Count (TPC)} = 13$) than when using only the low-level sub-model ($\text{TPC} = 11$). In other words, EnTri incorrectly predicts the auditorium as a movie theater $2$ times, compared to $4$ msiclassifications with the low-level sub-model alone. This suggests that the mid- and high-level sub-models, which we refer to as object-level sub-models, improve the recognition of auditoriums in two instances. As shown in the top row of~\Cref{discussion}(a), one of the auditorium samples that was incorrectly predicted as a movie theater by the low-level sub-model is provided. This error may be attributed to the similarity of the environment or background between these locations. However, our object-level sub-models utilize scene objects to differentiate between the two, as an auditorium typically has a stage and a curtain, while a movie theater typically has a screen. The confusion matrix also reveals that EnTri has a true positive count of $17$ for "waiting room", and didn't misclassify any as similar places such as "office" compared to the low-level sub-model that misclassified $2$ samples as "office". An example of this misclassification is provided in the second row of~\Cref{discussion}(a), where the object-level models focus on discriminative objects such as computer monitors, keyboards, and mice to accurately predict the "waiting room" scenes. Additionally, the "waiting room" has a greater number of chairs with respect to tables in comparison to the lobby scenes, which typically have an equal number of chairs and tables. These objects and their quantity can enhance the performance of object-level sub-models and make them effective in improving the accuracy of our scene recognition system.


        According to the confusion matrix~\Cref{LO-confusion}, it is also evident that the object-level sub-models exhibit inferior performance compared to the low-level sub-model and EnTri when it comes to predicting scenes with different categories, which include identical objects with approximately the same statistics, yet different in terms of environmental texture, lighting, and background. For example, EnTri achieved a higher true positive count of $22$ for the "inside bus" compared to when only using object-level sub-models, which had true positive counts of $21$. The object-level sub-models also misclassified an inside bus as an inside subway $4$ times, whereas EnTri reduced this misclassification to only $1$ instance. This implies that low-level features such as texture and lighting significantly improve the recognition of scenes inside a bus in three cases. As shown in the top row of~\Cref{discussion}(b), one of the samples that was incorrectly predicted by the object-level models is provided. The object-level models made inaccurate predictions of its category, as they relied on objects such as chairs alone, which are present both inside a bus and a subway and do not provide sufficient differentiation. In contrast, our low-level sub-model and EnTri correctly predicted the scene as being inside a bus. This was achieved by focusing on details of the environment and background texture, such as the colors and texture of the chairs and lighting in the bus. Similarly, in misclassifying "train station" as "subway", the rate of incorrect predictions for EnTri was half that of the low-level sub-model, with only $2$ errors compared to $4$ by the object-level sub-models. The bottom row of~\Cref{discussion}(b) illustrates one of these misclassifications. Since subway stations are typically located underground and have a darker environment and different background compared to train stations, low-level details such as color and lighting texture facilitated the prediction.
         
        From the results presented, we can conclude that high-level scene information, such as objects and their quantities, can be useful in improving scene recognition performance. Likewise, low-level information is useful in differentiating between scenes based on the textural details of the background and environment. Mid-level features serve as a bridge between these two by capturing both objects and their spatial layout. These different types of information complement each other to create a more representative feature space for the scene, leading to improved recognition performance.

\section{Conclusion}
\label{conclusion}
In this work, we have proposed EnTri, an ensemble learning framework for explainable scene recognition that employs three levels of perceptual features during the training process to enhance its accuracy and interpretability. The three core aspects of our work are demonstrated by (1) ensemble learning: We designed three sub-models, each intended to process a specific feature, to utilize ensemble learning for making predictions based on three levels of features. These sub-models incorporate multiple classifiers with different architectures to complement each other's strengths and weaknesses, promoting the effectiveness of ensemble learning in accuracy and generalization. (2) Interpretability: We developed an extension algorithm that provides visual and textual explanations so as to facilitate interpretability. The explanations highlight the scene properties, such as object categories and frequencies, object locations, and textural information, that contributed to the final prediction of the scene category, as well as confidence scores. Textual explanations offer notable benefits compared to heatmaps, particularly regarding the certainty with which non-expert users are able to interpret them as well as their ability to facilitate the identification of pivotal components and regions throughout the prediction process. By providing both visual and textual explanations, we enhance the interpretability and understanding of the prediction process. (3) Three-level representations: We leverage three levels of complementary feature representations, including pixel-level, semantic segmentation level, and object and frequency level, which alleviates the challenges posed by inter-class similarity and intra-class variation and leads to improvement in prediction accuracy. Such modular design also facilitates interpretability by dividing the problem into sub-problems and using multiple sub-models to solve them, making it easier to diagnose and optimize system behavior. Our model has been demonstrated to achieve competitive performance compared to state-of-the-art recognition accuracy on MIT67, SUN397, and UIUC8 datasets and effectively leverages key image features to generate intuitive visual and textual explanations to convey the reasoning behind its predictions.

     For future work, one possible direction is to choose a dataset for scene recognition that either has object detection and semantic segmentation annotations or create these annotations for datasets that do not have them and build mid-level and high-level representations by fine-tuning or training models explicitly on these annotations. These annotations can provide valuable supervision, potentially leading to more semantically meaningful representations that can promote better recognition performance.
     As our model has interpretability and multi-level features at its core, another possible direction is to extend our framework to high-stakes domains such as medical image classification, where understanding and trusting the reasoning of the system's decision-making becomes an absolute necessity.

\backmatter

\bmhead{Acknowledgements}

The authors would like to express their deepest gratitude to Quattoni, Xiao, and Fei-Fei Li for their remarkable contribution in providing the MIT67 dataset, SUN397 dataset, and UIUC8 dataset.

\section*{Declarations}

The authors declare that they have no known competing financial interests or personal relationships that could have appeared to influence the work reported in this paper.

\section*{Data availability}
All the data used in this work is from open and publicly available datasets for scene classification.


\bibliography{sn-bibliography}


\begin{thebibliography}{78}
\ifx \bisbn   \undefined \def \bisbn  #1{ISBN #1}\fi
\ifx \binits  \undefined \def \binits#1{#1}\fi
\ifx \bauthor  \undefined \def \bauthor#1{#1}\fi
\ifx \batitle  \undefined \def \batitle#1{#1}\fi
\ifx \bjtitle  \undefined \def \bjtitle#1{#1}\fi
\ifx \bvolume  \undefined \def \bvolume#1{\textbf{#1}}\fi
\ifx \byear  \undefined \def \byear#1{#1}\fi
\ifx \bissue  \undefined \def \bissue#1{#1}\fi
\ifx \bfpage  \undefined \def \bfpage#1{#1}\fi
\ifx \blpage  \undefined \def \blpage #1{#1}\fi
\ifx \burl  \undefined \def \burl#1{\textsf{#1}}\fi
\ifx \doiurl  \undefined \def \doiurl#1{\url{https://doi.org/#1}}\fi
\ifx \betal  \undefined \def \betal{\textit{et al.}}\fi
\ifx \binstitute  \undefined \def \binstitute#1{#1}\fi
\ifx \binstitutionaled  \undefined \def \binstitutionaled#1{#1}\fi
\ifx \bctitle  \undefined \def \bctitle#1{#1}\fi
\ifx \beditor  \undefined \def \beditor#1{#1}\fi
\ifx \bpublisher  \undefined \def \bpublisher#1{#1}\fi
\ifx \bbtitle  \undefined \def \bbtitle#1{#1}\fi
\ifx \bedition  \undefined \def \bedition#1{#1}\fi
\ifx \bseriesno  \undefined \def \bseriesno#1{#1}\fi
\ifx \blocation  \undefined \def \blocation#1{#1}\fi
\ifx \bsertitle  \undefined \def \bsertitle#1{#1}\fi
\ifx \bsnm \undefined \def \bsnm#1{#1}\fi
\ifx \bsuffix \undefined \def \bsuffix#1{#1}\fi
\ifx \bparticle \undefined \def \bparticle#1{#1}\fi
\ifx \barticle \undefined \def \barticle#1{#1}\fi
\bibcommenthead
\ifx \bconfdate \undefined \def \bconfdate #1{#1}\fi
\ifx \botherref \undefined \def \botherref #1{#1}\fi
\ifx \url \undefined \def \url#1{\textsf{#1}}\fi
\ifx \bchapter \undefined \def \bchapter#1{#1}\fi
\ifx \bbook \undefined \def \bbook#1{#1}\fi
\ifx \bcomment \undefined \def \bcomment#1{#1}\fi
\ifx \oauthor \undefined \def \oauthor#1{#1}\fi
\ifx \citeauthoryear \undefined \def \citeauthoryear#1{#1}\fi
\ifx \endbibitem  \undefined \def \endbibitem {}\fi
\ifx \bconflocation  \undefined \def \bconflocation#1{#1}\fi
\ifx \arxivurl  \undefined \def \arxivurl#1{\textsf{#1}}\fi
\csname PreBibitemsHook\endcsname

\bibitem[\protect\citeauthoryear{Xiao et~al.}{2010}]{xiao2010sun}
\begin{bchapter}
\bauthor{\bsnm{Xiao}, \binits{J.}},
\bauthor{\bsnm{Hays}, \binits{J.}},
\bauthor{\bsnm{Ehinger}, \binits{K.A.}},
\bauthor{\bsnm{Oliva}, \binits{A.}},
\bauthor{\bsnm{Torralba}, \binits{A.}}:
\bctitle{Sun database: Large-scale scene recognition from abbey to zoo}.
In: \bbtitle{2010 IEEE Computer Society Conference on Computer Vision and Pattern Recognition},
pp. \bfpage{3485}--\blpage{3492}
(\byear{2010}).
\bcomment{IEEE}
\end{bchapter}
\endbibitem

\bibitem[\protect\citeauthoryear{Li et~al.}{2020}]{li2020text}
\begin{barticle}
\bauthor{\bsnm{Li}, \binits{P.}},
\bauthor{\bsnm{Li}, \binits{X.}},
\bauthor{\bsnm{Pan}, \binits{H.}},
\bauthor{\bsnm{Khyam}, \binits{M.O.}},
\bauthor{\bsnm{Noor-A-Rahim}, \binits{M.}}:
\batitle{Text-based indoor place recognition with deep neural network}.
\bjtitle{Neurocomputing}
\bvolume{390},
\bfpage{239}--\blpage{247}
(\byear{2020})
\end{barticle}
\endbibitem

\bibitem[\protect\citeauthoryear{Wang et~al.}{2021}]{wang2021robust}
\begin{barticle}
\bauthor{\bsnm{Wang}, \binits{C.}},
\bauthor{\bsnm{Peng}, \binits{G.}},
\bauthor{\bsnm{Lin}, \binits{W.}}:
\batitle{Robust local metric learning via least square regression regularization for scene recognition}.
\bjtitle{Neurocomputing}
\bvolume{423},
\bfpage{179}--\blpage{189}
(\byear{2021})
\end{barticle}
\endbibitem

\bibitem[\protect\citeauthoryear{Sitaula et~al.}{2021}]{sitaula2021content}
\begin{barticle}
\bauthor{\bsnm{Sitaula}, \binits{C.}},
\bauthor{\bsnm{Aryal}, \binits{S.}},
\bauthor{\bsnm{Xiang}, \binits{Y.}},
\bauthor{\bsnm{Basnet}, \binits{A.}},
\bauthor{\bsnm{Lu}, \binits{X.}}:
\batitle{Content and context features for scene image representation}.
\bjtitle{Knowledge-Based Systems}
\bvolume{232},
\bfpage{107470}
(\byear{2021})
\end{barticle}
\endbibitem

\bibitem[\protect\citeauthoryear{Fan et~al.}{2022}]{fan2022indoor}
\begin{bchapter}
\bauthor{\bsnm{Fan}, \binits{X.}},
\bauthor{\bsnm{Zhu}, \binits{B.}},
\bauthor{\bsnm{Gao}, \binits{X.}},
\bauthor{\bsnm{Wang}, \binits{B.}},
\bauthor{\bsnm{Wang}, \binits{C.}},
\bauthor{\bsnm{Xu}, \binits{G.}}:
\bctitle{Indoor scene classification algorithm based on an object vector for robot applications}.
In: \bbtitle{2022 The 3rd International Conference on Artificial Intelligence in Electronics Engineering},
pp. \bfpage{55}--\blpage{64}
(\byear{2022})
\end{bchapter}
\endbibitem

\bibitem[\protect\citeauthoryear{Oliva and Torralba}{2001}]{oliva2001modeling}
\begin{barticle}
\bauthor{\bsnm{Oliva}, \binits{A.}},
\bauthor{\bsnm{Torralba}, \binits{A.}}:
\batitle{Modeling the shape of the scene: A holistic representation of the spatial envelope}.
\bjtitle{International journal of computer vision}
\bvolume{42},
\bfpage{145}--\blpage{175}
(\byear{2001})
\end{barticle}
\endbibitem

\bibitem[\protect\citeauthoryear{Lowe}{2004}]{lowe2004distinctive}
\begin{barticle}
\bauthor{\bsnm{Lowe}, \binits{D.G.}}:
\batitle{Distinctive image features from scale-invariant keypoints}.
\bjtitle{International journal of computer vision}
\bvolume{60},
\bfpage{91}--\blpage{110}
(\byear{2004})
\end{barticle}
\endbibitem

\bibitem[\protect\citeauthoryear{Csurka et~al.}{2004}]{csurka2004visual}
\begin{bchapter}
\bauthor{\bsnm{Csurka}, \binits{G.}},
\bauthor{\bsnm{Dance}, \binits{C.}},
\bauthor{\bsnm{Fan}, \binits{L.}},
\bauthor{\bsnm{Willamowski}, \binits{J.}},
\bauthor{\bsnm{Bray}, \binits{C.}}:
\bctitle{Visual categorization with bags of keypoints}.
In: \bbtitle{Workshop on Statistical Learning in Computer Vision, ECCV},
vol. \bseriesno{1},
pp. \bfpage{1}--\blpage{2}
(\byear{2004}).
\bcomment{Prague}
\end{bchapter}
\endbibitem

\bibitem[\protect\citeauthoryear{Li et~al.}{2010}]{li2010object}
\begin{botherref}
\oauthor{\bsnm{Li}, \binits{L.-J.}},
\oauthor{\bsnm{Su}, \binits{H.}},
\oauthor{\bsnm{Fei-Fei}, \binits{L.}},
\oauthor{\bsnm{Xing}, \binits{E.}}:
Object bank: A high-level image representation for scene classification \& semantic feature sparsification.
Advances in neural information processing systems
\textbf{23}
(2010)
\end{botherref}
\endbibitem

\bibitem[\protect\citeauthoryear{Voulodimos et~al.}{2018}]{voulodimos2018deep}
\begin{botherref}
\oauthor{\bsnm{Voulodimos}, \binits{A.}},
\oauthor{\bsnm{Doulamis}, \binits{N.}},
\oauthor{\bsnm{Doulamis}, \binits{A.}},
\oauthor{\bsnm{Protopapadakis}, \binits{E.}}:
Deep learning for computer vision: A brief review.
Computational intelligence and neuroscience
\textbf{2018}
(2018)
\end{botherref}
\endbibitem

\bibitem[\protect\citeauthoryear{Ragusa et~al.}{2019}]{ragusa2019survey}
\begin{barticle}
\bauthor{\bsnm{Ragusa}, \binits{E.}},
\bauthor{\bsnm{Cambria}, \binits{E.}},
\bauthor{\bsnm{Zunino}, \binits{R.}},
\bauthor{\bsnm{Gastaldo}, \binits{P.}}:
\batitle{A survey on deep learning in image polarity detection: Balancing generalization performances and computational costs}.
\bjtitle{Electronics}
\bvolume{8}(\bissue{7}),
\bfpage{783}
(\byear{2019})
\end{barticle}
\endbibitem

\bibitem[\protect\citeauthoryear{Machado et~al.}{2021}]{machado2021adversarial}
\begin{barticle}
\bauthor{\bsnm{Machado}, \binits{G.R.}},
\bauthor{\bsnm{Silva}, \binits{E.}},
\bauthor{\bsnm{Goldschmidt}, \binits{R.R.}}:
\batitle{Adversarial machine learning in image classification: A survey toward the defender’s perspective}.
\bjtitle{ACM Computing Surveys (CSUR)}
\bvolume{55}(\bissue{1}),
\bfpage{1}--\blpage{38}
(\byear{2021})
\end{barticle}
\endbibitem

\bibitem[\protect\citeauthoryear{Krizhevsky et~al.}{2017}]{krizhevsky2017imagenet}
\begin{barticle}
\bauthor{\bsnm{Krizhevsky}, \binits{A.}},
\bauthor{\bsnm{Sutskever}, \binits{I.}},
\bauthor{\bsnm{Hinton}, \binits{G.E.}}:
\batitle{Imagenet classification with deep convolutional neural networks}.
\bjtitle{Communications of the ACM}
\bvolume{60}(\bissue{6}),
\bfpage{84}--\blpage{90}
(\byear{2017})
\end{barticle}
\endbibitem

\bibitem[\protect\citeauthoryear{Simonyan and Zisserman}{2014}]{simonyan2014very}
\begin{botherref}
\oauthor{\bsnm{Simonyan}, \binits{K.}},
\oauthor{\bsnm{Zisserman}, \binits{A.}}:
Very deep convolutional networks for large-scale image recognition.
arXiv preprint arXiv:1409.1556
(2014)
\end{botherref}
\endbibitem

\bibitem[\protect\citeauthoryear{Szegedy et~al.}{2015}]{szegedy2015going}
\begin{bchapter}
\bauthor{\bsnm{Szegedy}, \binits{C.}},
\bauthor{\bsnm{Liu}, \binits{W.}},
\bauthor{\bsnm{Jia}, \binits{Y.}},
\bauthor{\bsnm{Sermanet}, \binits{P.}},
\bauthor{\bsnm{Reed}, \binits{S.}},
\bauthor{\bsnm{Anguelov}, \binits{D.}},
\bauthor{\bsnm{Erhan}, \binits{D.}},
\bauthor{\bsnm{Vanhoucke}, \binits{V.}},
\bauthor{\bsnm{Rabinovich}, \binits{A.}}:
\bctitle{Going deeper with convolutions}.
In: \bbtitle{Proceedings of the IEEE Conference on Computer Vision and Pattern Recognition},
pp. \bfpage{1}--\blpage{9}
(\byear{2015})
\end{bchapter}
\endbibitem

\bibitem[\protect\citeauthoryear{He et~al.}{2016}]{he2016deep}
\begin{bchapter}
\bauthor{\bsnm{He}, \binits{K.}},
\bauthor{\bsnm{Zhang}, \binits{X.}},
\bauthor{\bsnm{Ren}, \binits{S.}},
\bauthor{\bsnm{Sun}, \binits{J.}}:
\bctitle{Deep residual learning for image recognition}.
In: \bbtitle{Proceedings of the IEEE Conference on Computer Vision and Pattern Recognition},
pp. \bfpage{770}--\blpage{778}
(\byear{2016})
\end{bchapter}
\endbibitem

\bibitem[\protect\citeauthoryear{Huang et~al.}{2017}]{huang2017densely}
\begin{bchapter}
\bauthor{\bsnm{Huang}, \binits{G.}},
\bauthor{\bsnm{Liu}, \binits{Z.}},
\bauthor{\bsnm{Van Der~Maaten}, \binits{L.}},
\bauthor{\bsnm{Weinberger}, \binits{K.Q.}}:
\bctitle{Densely connected convolutional networks}.
In: \bbtitle{Proceedings of the IEEE Conference on Computer Vision and Pattern Recognition},
pp. \bfpage{4700}--\blpage{4708}
(\byear{2017})
\end{bchapter}
\endbibitem

\bibitem[\protect\citeauthoryear{LeCun et~al.}{2015}]{lecun2015deep}
\begin{barticle}
\bauthor{\bsnm{LeCun}, \binits{Y.}},
\bauthor{\bsnm{Bengio}, \binits{Y.}},
\bauthor{\bsnm{Hinton}, \binits{G.}}:
\batitle{Deep learning}.
\bjtitle{nature}
\bvolume{521}(\bissue{7553}),
\bfpage{436}--\blpage{444}
(\byear{2015})
\end{barticle}
\endbibitem

\bibitem[\protect\citeauthoryear{He et~al.}{2019}]{he2019bag}
\begin{bchapter}
\bauthor{\bsnm{He}, \binits{T.}},
\bauthor{\bsnm{Zhang}, \binits{Z.}},
\bauthor{\bsnm{Zhang}, \binits{H.}},
\bauthor{\bsnm{Zhang}, \binits{Z.}},
\bauthor{\bsnm{Xie}, \binits{J.}},
\bauthor{\bsnm{Li}, \binits{M.}}:
\bctitle{Bag of tricks for image classification with convolutional neural networks}.
In: \bbtitle{Proceedings of the IEEE/CVF Conference on Computer Vision and Pattern Recognition},
pp. \bfpage{558}--\blpage{567}
(\byear{2019})
\end{bchapter}
\endbibitem

\bibitem[\protect\citeauthoryear{Li et~al.}{2021}]{li2021survey}
\begin{botherref}
\oauthor{\bsnm{Li}, \binits{Z.}},
\oauthor{\bsnm{Liu}, \binits{F.}},
\oauthor{\bsnm{Yang}, \binits{W.}},
\oauthor{\bsnm{Peng}, \binits{S.}},
\oauthor{\bsnm{Zhou}, \binits{J.}}:
A survey of convolutional neural networks: analysis, applications, and prospects.
IEEE transactions on neural networks and learning systems
(2021)
\end{botherref}
\endbibitem

\bibitem[\protect\citeauthoryear{Cheng et~al.}{2018}]{cheng2018scene}
\begin{barticle}
\bauthor{\bsnm{Cheng}, \binits{X.}},
\bauthor{\bsnm{Lu}, \binits{J.}},
\bauthor{\bsnm{Feng}, \binits{J.}},
\bauthor{\bsnm{Yuan}, \binits{B.}},
\bauthor{\bsnm{Zhou}, \binits{J.}}:
\batitle{Scene recognition with objectness}.
\bjtitle{Pattern Recognition}
\bvolume{74},
\bfpage{474}--\blpage{487}
(\byear{2018})
\end{barticle}
\endbibitem

\bibitem[\protect\citeauthoryear{Xiong et~al.}{2020}]{xiong2020msn}
\begin{barticle}
\bauthor{\bsnm{Xiong}, \binits{Z.}},
\bauthor{\bsnm{Yuan}, \binits{Y.}},
\bauthor{\bsnm{Wang}, \binits{Q.}}:
\batitle{Msn: Modality separation networks for rgb-d scene recognition}.
\bjtitle{Neurocomputing}
\bvolume{373},
\bfpage{81}--\blpage{89}
(\byear{2020})
\end{barticle}
\endbibitem

\bibitem[\protect\citeauthoryear{Lin et~al.}{2022}]{lin2022scene}
\begin{barticle}
\bauthor{\bsnm{Lin}, \binits{C.}},
\bauthor{\bsnm{Lee}, \binits{F.}},
\bauthor{\bsnm{Xie}, \binits{L.}},
\bauthor{\bsnm{Cai}, \binits{J.}},
\bauthor{\bsnm{Chen}, \binits{H.}},
\bauthor{\bsnm{Liu}, \binits{L.}},
\bauthor{\bsnm{Chen}, \binits{Q.}}:
\batitle{Scene recognition using multiple representation network}.
\bjtitle{Applied Soft Computing}
\bvolume{118},
\bfpage{108530}
(\byear{2022})
\end{barticle}
\endbibitem

\bibitem[\protect\citeauthoryear{Nanni and Lumini}{2013}]{nanni2013heterogeneous}
\begin{barticle}
\bauthor{\bsnm{Nanni}, \binits{L.}},
\bauthor{\bsnm{Lumini}, \binits{A.}}:
\batitle{Heterogeneous bag-of-features for object/scene recognition}.
\bjtitle{Applied Soft Computing}
\bvolume{13}(\bissue{4}),
\bfpage{2171}--\blpage{2178}
(\byear{2013})
\end{barticle}
\endbibitem

\bibitem[\protect\citeauthoryear{Bai and Tang}{2018}]{bai2018softly}
\begin{barticle}
\bauthor{\bsnm{Bai}, \binits{S.}},
\bauthor{\bsnm{Tang}, \binits{H.}}:
\batitle{Softly combining an ensemble of classifiers learned from a single convolutional neural network for scene categorization}.
\bjtitle{Applied Soft Computing}
\bvolume{67},
\bfpage{183}--\blpage{196}
(\byear{2018})
\end{barticle}
\endbibitem

\bibitem[\protect\citeauthoryear{Hernandez et~al.}{2019}]{hernandez2019indoor}
\begin{bchapter}
\bauthor{\bsnm{Hernandez}, \binits{A.C.}},
\bauthor{\bsnm{Gomez}, \binits{C.}},
\bauthor{\bsnm{Derner}, \binits{E.}},
\bauthor{\bsnm{Barber}, \binits{R.}}:
\bctitle{Indoor scene recognition based on weighted voting schemes}.
In: \bbtitle{2019 European Conference on Mobile Robots (ECMR)},
pp. \bfpage{1}--\blpage{6}
(\byear{2019}).
\bcomment{IEEE}
\end{bchapter}
\endbibitem

\bibitem[\protect\citeauthoryear{Ribeiro et~al.}{2016}]{ribeiro2016should}
\begin{bchapter}
\bauthor{\bsnm{Ribeiro}, \binits{M.T.}},
\bauthor{\bsnm{Singh}, \binits{S.}},
\bauthor{\bsnm{Guestrin}, \binits{C.}}:
\bctitle{" why should i trust you?" explaining the predictions of any classifier}.
In: \bbtitle{Proceedings of the 22nd ACM SIGKDD International Conference on Knowledge Discovery and Data Mining},
pp. \bfpage{1135}--\blpage{1144}
(\byear{2016})
\end{bchapter}
\endbibitem

\bibitem[\protect\citeauthoryear{Do{\v{s}}ilovi{\'c} et~al.}{2018}]{dovsilovic2018explainable}
\begin{bchapter}
\bauthor{\bsnm{Do{\v{s}}ilovi{\'c}}, \binits{F.K.}},
\bauthor{\bsnm{Br{\v{c}}i{\'c}}, \binits{M.}},
\bauthor{\bsnm{Hlupi{\'c}}, \binits{N.}}:
\bctitle{Explainable artificial intelligence: A survey}.
In: \bbtitle{2018 41st International Convention on Information and Communication Technology, Electronics and Microelectronics (MIPRO)},
pp. \bfpage{0210}--\blpage{0215}
(\byear{2018}).
\bcomment{IEEE}
\end{bchapter}
\endbibitem

\bibitem[\protect\citeauthoryear{Islam et~al.}{2021}]{islam2021explainable}
\begin{botherref}
\oauthor{\bsnm{Islam}, \binits{S.R.}},
\oauthor{\bsnm{Eberle}, \binits{W.}},
\oauthor{\bsnm{Ghafoor}, \binits{S.K.}},
\oauthor{\bsnm{Ahmed}, \binits{M.}}:
Explainable artificial intelligence approaches: A survey.
arXiv preprint arXiv:2101.09429
(2021)
\end{botherref}
\endbibitem

\bibitem[\protect\citeauthoryear{Miller}{2019}]{miller2019explanation}
\begin{barticle}
\bauthor{\bsnm{Miller}, \binits{T.}}:
\batitle{Explanation in artificial intelligence: Insights from the social sciences}.
\bjtitle{Artificial intelligence}
\bvolume{267},
\bfpage{1}--\blpage{38}
(\byear{2019})
\end{barticle}
\endbibitem

\bibitem[\protect\citeauthoryear{Das and Rad}{2020}]{das2020opportunities}
\begin{botherref}
\oauthor{\bsnm{Das}, \binits{A.}},
\oauthor{\bsnm{Rad}, \binits{P.}}:
Opportunities and challenges in explainable artificial intelligence (xai): A survey.
arXiv preprint arXiv:2006.11371
(2020)
\end{botherref}
\endbibitem

\bibitem[\protect\citeauthoryear{Zhang et~al.}{2022}]{zhang2022explainable}
\begin{barticle}
\bauthor{\bsnm{Zhang}, \binits{X.}},
\bauthor{\bsnm{Chan}, \binits{F.T.}},
\bauthor{\bsnm{Mahadevan}, \binits{S.}}:
\batitle{Explainable machine learning in image classification models: An uncertainty quantification perspective}.
\bjtitle{Knowledge-Based Systems}
\bvolume{243},
\bfpage{108418}
(\byear{2022})
\end{barticle}
\endbibitem

\bibitem[\protect\citeauthoryear{Tjoa and Guan}{2020}]{tjoa2020survey}
\begin{barticle}
\bauthor{\bsnm{Tjoa}, \binits{E.}},
\bauthor{\bsnm{Guan}, \binits{C.}}:
\batitle{A survey on explainable artificial intelligence (xai): Toward medical xai}.
\bjtitle{IEEE transactions on neural networks and learning systems}
\bvolume{32}(\bissue{11}),
\bfpage{4793}--\blpage{4813}
(\byear{2020})
\end{barticle}
\endbibitem

\bibitem[\protect\citeauthoryear{Selvaraju et~al.}{2017}]{selvaraju2017grad}
\begin{bchapter}
\bauthor{\bsnm{Selvaraju}, \binits{R.R.}},
\bauthor{\bsnm{Cogswell}, \binits{M.}},
\bauthor{\bsnm{Das}, \binits{A.}},
\bauthor{\bsnm{Vedantam}, \binits{R.}},
\bauthor{\bsnm{Parikh}, \binits{D.}},
\bauthor{\bsnm{Batra}, \binits{D.}}:
\bctitle{Grad-cam: Visual explanations from deep networks via gradient-based localization}.
In: \bbtitle{Proceedings of the IEEE International Conference on Computer Vision},
pp. \bfpage{618}--\blpage{626}
(\byear{2017})
\end{bchapter}
\endbibitem

\bibitem[\protect\citeauthoryear{Yang and Ramanan}{2015}]{yang2015multi}
\begin{bchapter}
\bauthor{\bsnm{Yang}, \binits{S.}},
\bauthor{\bsnm{Ramanan}, \binits{D.}}:
\bctitle{Multi-scale recognition with dag-cnns}.
In: \bbtitle{Proceedings of the IEEE International Conference on Computer Vision},
pp. \bfpage{1215}--\blpage{1223}
(\byear{2015})
\end{bchapter}
\endbibitem

\bibitem[\protect\citeauthoryear{Lin et~al.}{2017}]{lin2017feature}
\begin{bchapter}
\bauthor{\bsnm{Lin}, \binits{T.-Y.}},
\bauthor{\bsnm{Doll{\'a}r}, \binits{P.}},
\bauthor{\bsnm{Girshick}, \binits{R.}},
\bauthor{\bsnm{He}, \binits{K.}},
\bauthor{\bsnm{Hariharan}, \binits{B.}},
\bauthor{\bsnm{Belongie}, \binits{S.}}:
\bctitle{Feature pyramid networks for object detection}.
In: \bbtitle{Proceedings of the IEEE Conference on Computer Vision and Pattern Recognition},
pp. \bfpage{2117}--\blpage{2125}
(\byear{2017})
\end{bchapter}
\endbibitem

\bibitem[\protect\citeauthoryear{Quattoni and Torralba}{2009}]{quattoni2009recognizing}
\begin{bchapter}
\bauthor{\bsnm{Quattoni}, \binits{A.}},
\bauthor{\bsnm{Torralba}, \binits{A.}}:
\bctitle{Recognizing indoor scenes}.
In: \bbtitle{2009 IEEE Conference on Computer Vision and Pattern Recognition},
pp. \bfpage{413}--\blpage{420}
(\byear{2009}).
\bcomment{IEEE}
\end{bchapter}
\endbibitem

\bibitem[\protect\citeauthoryear{Li and Fei-Fei}{2007}]{li2007and}
\begin{bchapter}
\bauthor{\bsnm{Li}, \binits{L.-J.}},
\bauthor{\bsnm{Fei-Fei}, \binits{L.}}:
\bctitle{What, where and who? classifying events by scene and object recognition}.
In: \bbtitle{2007 IEEE 11th International Conference on Computer Vision},
pp. \bfpage{1}--\blpage{8}
(\byear{2007}).
\bcomment{IEEE}
\end{bchapter}
\endbibitem

\bibitem[\protect\citeauthoryear{Lazebnik et~al.}{2006}]{lazebnik2006beyond}
\begin{bchapter}
\bauthor{\bsnm{Lazebnik}, \binits{S.}},
\bauthor{\bsnm{Schmid}, \binits{C.}},
\bauthor{\bsnm{Ponce}, \binits{J.}}:
\bctitle{Beyond bags of features: Spatial pyramid matching for recognizing natural scene categories}.
In: \bbtitle{2006 IEEE Computer Society Conference on Computer Vision and Pattern Recognition (CVPR'06)},
vol. \bseriesno{2},
pp. \bfpage{2169}--\blpage{2178}
(\byear{2006}).
\bcomment{IEEE}
\end{bchapter}
\endbibitem

\bibitem[\protect\citeauthoryear{Wu and Rehg}{2010}]{wu2010centrist}
\begin{barticle}
\bauthor{\bsnm{Wu}, \binits{J.}},
\bauthor{\bsnm{Rehg}, \binits{J.M.}}:
\batitle{Centrist: A visual descriptor for scene categorization}.
\bjtitle{IEEE transactions on pattern analysis and machine intelligence}
\bvolume{33}(\bissue{8}),
\bfpage{1489}--\blpage{1501}
(\byear{2010})
\end{barticle}
\endbibitem

\bibitem[\protect\citeauthoryear{Qian et~al.}{2018}]{qian2018learning}
\begin{bchapter}
\bauthor{\bsnm{Qian}, \binits{C.}},
\bauthor{\bsnm{Chaturvedi}, \binits{I.}},
\bauthor{\bsnm{Poria}, \binits{S.}},
\bauthor{\bsnm{Cambria}, \binits{E.}},
\bauthor{\bsnm{Malandri}, \binits{L.}}:
\bctitle{Learning visual concepts in images using temporal convolutional networks}.
In: \bbtitle{2018 IEEE Symposium Series on Computational Intelligence (SSCI)},
pp. \bfpage{1280}--\blpage{1284}
(\byear{2018}).
\bcomment{IEEE}
\end{bchapter}
\endbibitem

\bibitem[\protect\citeauthoryear{Liu et~al.}{2022}]{liu2022convnet}
\begin{bchapter}
\bauthor{\bsnm{Liu}, \binits{Z.}},
\bauthor{\bsnm{Mao}, \binits{H.}},
\bauthor{\bsnm{Wu}, \binits{C.-Y.}},
\bauthor{\bsnm{Feichtenhofer}, \binits{C.}},
\bauthor{\bsnm{Darrell}, \binits{T.}},
\bauthor{\bsnm{Xie}, \binits{S.}}:
\bctitle{A convnet for the 2020s}.
In: \bbtitle{Proceedings of the IEEE/CVF Conference on Computer Vision and Pattern Recognition},
pp. \bfpage{11976}--\blpage{11986}
(\byear{2022})
\end{bchapter}
\endbibitem

\bibitem[\protect\citeauthoryear{Gong et~al.}{2014}]{gong2014multi}
\begin{bchapter}
\bauthor{\bsnm{Gong}, \binits{Y.}},
\bauthor{\bsnm{Wang}, \binits{L.}},
\bauthor{\bsnm{Guo}, \binits{R.}},
\bauthor{\bsnm{Lazebnik}, \binits{S.}}:
\bctitle{Multi-scale orderless pooling of deep convolutional activation features}.
In: \bbtitle{Computer Vision--ECCV 2014: 13th European Conference, Zurich, Switzerland, September 6-12, 2014, Proceedings, Part VII 13},
pp. \bfpage{392}--\blpage{407}
(\byear{2014}).
\bcomment{Springer}
\end{bchapter}
\endbibitem

\bibitem[\protect\citeauthoryear{Herranz et~al.}{2016}]{herranz2016scene}
\begin{bchapter}
\bauthor{\bsnm{Herranz}, \binits{L.}},
\bauthor{\bsnm{Jiang}, \binits{S.}},
\bauthor{\bsnm{Li}, \binits{X.}}:
\bctitle{Scene recognition with cnns: objects, scales and dataset bias}.
In: \bbtitle{Proceedings of the IEEE Conference on Computer Vision and Pattern Recognition},
pp. \bfpage{571}--\blpage{579}
(\byear{2016})
\end{bchapter}
\endbibitem

\bibitem[\protect\citeauthoryear{Xie et~al.}{2017}]{xie2017lg}
\begin{barticle}
\bauthor{\bsnm{Xie}, \binits{G.-S.}},
\bauthor{\bsnm{Zhang}, \binits{X.-Y.}},
\bauthor{\bsnm{Yang}, \binits{W.}},
\bauthor{\bsnm{Xu}, \binits{M.}},
\bauthor{\bsnm{Yan}, \binits{S.}},
\bauthor{\bsnm{Liu}, \binits{C.-L.}}:
\batitle{Lg-cnn: From local parts to global discrimination for fine-grained recognition}.
\bjtitle{Pattern Recognition}
\bvolume{71},
\bfpage{118}--\blpage{131}
(\byear{2017})
\end{barticle}
\endbibitem

\bibitem[\protect\citeauthoryear{Wang et~al.}{2017}]{wang2017weakly}
\begin{barticle}
\bauthor{\bsnm{Wang}, \binits{Z.}},
\bauthor{\bsnm{Wang}, \binits{L.}},
\bauthor{\bsnm{Wang}, \binits{Y.}},
\bauthor{\bsnm{Zhang}, \binits{B.}},
\bauthor{\bsnm{Qiao}, \binits{Y.}}:
\batitle{Weakly supervised patchnets: Describing and aggregating local patches for scene recognition}.
\bjtitle{IEEE Transactions on Image Processing}
\bvolume{26}(\bissue{4}),
\bfpage{2028}--\blpage{2041}
(\byear{2017})
\end{barticle}
\endbibitem

\bibitem[\protect\citeauthoryear{L{\'o}pez-Cifuentes et~al.}{2020}]{lopez2020semantic}
\begin{barticle}
\bauthor{\bsnm{L{\'o}pez-Cifuentes}, \binits{A.}},
\bauthor{\bsnm{Escudero-Vinolo}, \binits{M.}},
\bauthor{\bsnm{Besc{\'o}s}, \binits{J.}},
\bauthor{\bsnm{Garc{\'\i}a-Mart{\'\i}n}, \binits{{\'A}.}}:
\batitle{Semantic-aware scene recognition}.
\bjtitle{Pattern Recognition}
\bvolume{102},
\bfpage{107256}
(\byear{2020})
\end{barticle}
\endbibitem

\bibitem[\protect\citeauthoryear{Chen et~al.}{2019}]{chen2019scene}
\begin{botherref}
\oauthor{\bsnm{Chen}, \binits{B.X.}},
\oauthor{\bsnm{Sahdev}, \binits{R.}},
\oauthor{\bsnm{Wu}, \binits{D.}},
\oauthor{\bsnm{Zhao}, \binits{X.}},
\oauthor{\bsnm{Papagelis}, \binits{M.}},
\oauthor{\bsnm{Tsotsos}, \binits{J.K.}}:
Scene classification in indoor environments for robots using context based word embeddings.
arXiv preprint arXiv:1908.06422
(2019)
\end{botherref}
\endbibitem

\bibitem[\protect\citeauthoryear{Heikel and Espinosa-Leal}{2022}]{heikel2022indoor}
\begin{barticle}
\bauthor{\bsnm{Heikel}, \binits{E.}},
\bauthor{\bsnm{Espinosa-Leal}, \binits{L.}}:
\batitle{Indoor scene recognition via object detection and tf-idf}.
\bjtitle{Journal of Imaging}
\bvolume{8}(\bissue{8}),
\bfpage{209}
(\byear{2022})
\end{barticle}
\endbibitem

\bibitem[\protect\citeauthoryear{Redmon et~al.}{2016}]{redmon2016you}
\begin{bchapter}
\bauthor{\bsnm{Redmon}, \binits{J.}},
\bauthor{\bsnm{Divvala}, \binits{S.}},
\bauthor{\bsnm{Girshick}, \binits{R.}},
\bauthor{\bsnm{Farhadi}, \binits{A.}}:
\bctitle{You only look once: Unified, real-time object detection}.
In: \bbtitle{Proceedings of the IEEE Conference on Computer Vision and Pattern Recognition},
pp. \bfpage{779}--\blpage{788}
(\byear{2016})
\end{bchapter}
\endbibitem

\bibitem[\protect\citeauthoryear{Wang et~al.}{2022}]{wang2022embedding}
\begin{barticle}
\bauthor{\bsnm{Wang}, \binits{C.}},
\bauthor{\bsnm{Peng}, \binits{G.}},
\bauthor{\bsnm{De~Baets}, \binits{B.}}:
\batitle{Embedding metric learning into an extreme learning machine for scene recognition}.
\bjtitle{Expert Systems with Applications}
\bvolume{203},
\bfpage{117505}
(\byear{2022})
\end{barticle}
\endbibitem

\bibitem[\protect\citeauthoryear{Zou et~al.}{2022}]{zou2022adanff}
\begin{barticle}
\bauthor{\bsnm{Zou}, \binits{Z.}},
\bauthor{\bsnm{Liu}, \binits{W.}},
\bauthor{\bsnm{Xing}, \binits{W.}}:
\batitle{Adanff: A new method for adaptive nonnegative multi-feature fusion to scene classification}.
\bjtitle{Pattern Recognition}
\bvolume{123},
\bfpage{108402}
(\byear{2022})
\end{barticle}
\endbibitem

\bibitem[\protect\citeauthoryear{Lundberg and Lee}{2017}]{lundberg2017unified}
\begin{botherref}
\oauthor{\bsnm{Lundberg}, \binits{S.M.}},
\oauthor{\bsnm{Lee}, \binits{S.-I.}}:
A unified approach to interpreting model predictions.
Advances in neural information processing systems
\textbf{30}
(2017)
\end{botherref}
\endbibitem

\bibitem[\protect\citeauthoryear{Bach et~al.}{2015}]{bach2015pixel}
\begin{barticle}
\bauthor{\bsnm{Bach}, \binits{S.}},
\bauthor{\bsnm{Binder}, \binits{A.}},
\bauthor{\bsnm{Montavon}, \binits{G.}},
\bauthor{\bsnm{Klauschen}, \binits{F.}},
\bauthor{\bsnm{M{\"u}ller}, \binits{K.-R.}},
\bauthor{\bsnm{Samek}, \binits{W.}}:
\batitle{On pixel-wise explanations for non-linear classifier decisions by layer-wise relevance propagation}.
\bjtitle{PloS one}
\bvolume{10}(\bissue{7}),
\bfpage{0130140}
(\byear{2015})
\end{barticle}
\endbibitem

\bibitem[\protect\citeauthoryear{Stefanini et~al.}{2022}]{stefanini2022show}
\begin{barticle}
\bauthor{\bsnm{Stefanini}, \binits{M.}},
\bauthor{\bsnm{Cornia}, \binits{M.}},
\bauthor{\bsnm{Baraldi}, \binits{L.}},
\bauthor{\bsnm{Cascianelli}, \binits{S.}},
\bauthor{\bsnm{Fiameni}, \binits{G.}},
\bauthor{\bsnm{Cucchiara}, \binits{R.}}:
\batitle{From show to tell: a survey on deep learning-based image captioning}.
\bjtitle{IEEE transactions on pattern analysis and machine intelligence}
\bvolume{45}(\bissue{1}),
\bfpage{539}--\blpage{559}
(\byear{2022})
\end{barticle}
\endbibitem

\bibitem[\protect\citeauthoryear{Vinyals et~al.}{2015}]{vinyals2015show}
\begin{bchapter}
\bauthor{\bsnm{Vinyals}, \binits{O.}},
\bauthor{\bsnm{Toshev}, \binits{A.}},
\bauthor{\bsnm{Bengio}, \binits{S.}},
\bauthor{\bsnm{Erhan}, \binits{D.}}:
\bctitle{Show and tell: A neural image caption generator}.
In: \bbtitle{Proceedings of the IEEE Conference on Computer Vision and Pattern Recognition},
pp. \bfpage{3156}--\blpage{3164}
(\byear{2015})
\end{bchapter}
\endbibitem

\bibitem[\protect\citeauthoryear{Huang et~al.}{2019}]{huang2019attention}
\begin{bchapter}
\bauthor{\bsnm{Huang}, \binits{L.}},
\bauthor{\bsnm{Wang}, \binits{W.}},
\bauthor{\bsnm{Chen}, \binits{J.}},
\bauthor{\bsnm{Wei}, \binits{X.-Y.}}:
\bctitle{Attention on attention for image captioning}.
In: \bbtitle{Proceedings of the IEEE/CVF International Conference on Computer Vision},
pp. \bfpage{4634}--\blpage{4643}
(\byear{2019})
\end{bchapter}
\endbibitem

\bibitem[\protect\citeauthoryear{Anjomshoae et~al.}{2021}]{anjomshoae2021context}
\begin{barticle}
\bauthor{\bsnm{Anjomshoae}, \binits{S.}},
\bauthor{\bsnm{Omeiza}, \binits{D.}},
\bauthor{\bsnm{Jiang}, \binits{L.}}:
\batitle{Context-based image explanations for deep neural networks}.
\bjtitle{Image and Vision Computing}
\bvolume{116},
\bfpage{104310}
(\byear{2021})
\end{barticle}
\endbibitem

\bibitem[\protect\citeauthoryear{Aminimehr et~al.}{}]{aminimehr4385953tbexplain}
\begin{botherref}
\oauthor{\bsnm{Aminimehr}, \binits{A.}},
\oauthor{\bsnm{Khani}, \binits{P.}},
\oauthor{\bsnm{Molaei}, \binits{A.}},
\oauthor{\bsnm{Kazemeini}, \binits{A.}},
\oauthor{\bsnm{Cambria}, \binits{E.}}:
Tbexplain: A text-based explanation method for scene classification models with the statistical prediction correction.
Available at SSRN 4385953
\end{botherref}
\endbibitem

\bibitem[\protect\citeauthoryear{Wolpert}{1992}]{wolpert1992stacked}
\begin{barticle}
\bauthor{\bsnm{Wolpert}, \binits{D.H.}}:
\batitle{Stacked generalization}.
\bjtitle{Neural networks}
\bvolume{5}(\bissue{2}),
\bfpage{241}--\blpage{259}
(\byear{1992})
\end{barticle}
\endbibitem

\bibitem[\protect\citeauthoryear{Tan and Le}{2021}]{tan2021efficientnetv2}
\begin{bchapter}
\bauthor{\bsnm{Tan}, \binits{M.}},
\bauthor{\bsnm{Le}, \binits{Q.}}:
\bctitle{Efficientnetv2: Smaller models and faster training}.
In: \bbtitle{International Conference on Machine Learning},
pp. \bfpage{10096}--\blpage{10106}
(\byear{2021}).
\bcomment{PMLR}
\end{bchapter}
\endbibitem

\bibitem[\protect\citeauthoryear{Deng et~al.}{2009}]{deng2009imagenet}
\begin{bchapter}
\bauthor{\bsnm{Deng}, \binits{J.}},
\bauthor{\bsnm{Dong}, \binits{W.}},
\bauthor{\bsnm{Socher}, \binits{R.}},
\bauthor{\bsnm{Li}, \binits{L.-J.}},
\bauthor{\bsnm{Li}, \binits{K.}},
\bauthor{\bsnm{Fei-Fei}, \binits{L.}}:
\bctitle{Imagenet: A large-scale hierarchical image database}.
In: \bbtitle{2009 IEEE Conference on Computer Vision and Pattern Recognition},
pp. \bfpage{248}--\blpage{255}
(\byear{2009}).
\bcomment{Ieee}
\end{bchapter}
\endbibitem

\bibitem[\protect\citeauthoryear{Chen et~al.}{2017}]{chen2017deeplab}
\begin{barticle}
\bauthor{\bsnm{Chen}, \binits{L.-C.}},
\bauthor{\bsnm{Papandreou}, \binits{G.}},
\bauthor{\bsnm{Kokkinos}, \binits{I.}},
\bauthor{\bsnm{Murphy}, \binits{K.}},
\bauthor{\bsnm{Yuille}, \binits{A.L.}}:
\batitle{Deeplab: Semantic image segmentation with deep convolutional nets, atrous convolution, and fully connected crfs}.
\bjtitle{IEEE transactions on pattern analysis and machine intelligence}
\bvolume{40}(\bissue{4}),
\bfpage{834}--\blpage{848}
(\byear{2017})
\end{barticle}
\endbibitem

\bibitem[\protect\citeauthoryear{Caesar et~al.}{2018}]{caesar2018coco}
\begin{bchapter}
\bauthor{\bsnm{Caesar}, \binits{H.}},
\bauthor{\bsnm{Uijlings}, \binits{J.}},
\bauthor{\bsnm{Ferrari}, \binits{V.}}:
\bctitle{Coco-stuff: Thing and stuff classes in context}.
In: \bbtitle{Proceedings of the IEEE Conference on Computer Vision and Pattern Recognition},
pp. \bfpage{1209}--\blpage{1218}
(\byear{2018})
\end{bchapter}
\endbibitem

\bibitem[\protect\citeauthoryear{Zhao et~al.}{2017}]{zhao2017pyramid}
\begin{bchapter}
\bauthor{\bsnm{Zhao}, \binits{H.}},
\bauthor{\bsnm{Shi}, \binits{J.}},
\bauthor{\bsnm{Qi}, \binits{X.}},
\bauthor{\bsnm{Wang}, \binits{X.}},
\bauthor{\bsnm{Jia}, \binits{J.}}:
\bctitle{Pyramid scene parsing network}.
In: \bbtitle{Proceedings of the IEEE Conference on Computer Vision and Pattern Recognition},
pp. \bfpage{2881}--\blpage{2890}
(\byear{2017})
\end{bchapter}
\endbibitem

\bibitem[\protect\citeauthoryear{Zhou et~al.}{2019}]{zhou2019semantic}
\begin{barticle}
\bauthor{\bsnm{Zhou}, \binits{B.}},
\bauthor{\bsnm{Zhao}, \binits{H.}},
\bauthor{\bsnm{Puig}, \binits{X.}},
\bauthor{\bsnm{Xiao}, \binits{T.}},
\bauthor{\bsnm{Fidler}, \binits{S.}},
\bauthor{\bsnm{Barriuso}, \binits{A.}},
\bauthor{\bsnm{Torralba}, \binits{A.}}:
\batitle{Semantic understanding of scenes through the ade20k dataset}.
\bjtitle{International Journal of Computer Vision}
\bvolume{127},
\bfpage{302}--\blpage{321}
(\byear{2019})
\end{barticle}
\endbibitem

\bibitem[\protect\citeauthoryear{Szegedy et~al.}{2017}]{szegedy2017inception}
\begin{bchapter}
\bauthor{\bsnm{Szegedy}, \binits{C.}},
\bauthor{\bsnm{Ioffe}, \binits{S.}},
\bauthor{\bsnm{Vanhoucke}, \binits{V.}},
\bauthor{\bsnm{Alemi}, \binits{A.}}:
\bctitle{Inception-v4, inception-resnet and the impact of residual connections on learning}.
In: \bbtitle{Proceedings of the AAAI Conference on Artificial Intelligence},
vol. \bseriesno{31}
(\byear{2017})
\end{bchapter}
\endbibitem

\bibitem[\protect\citeauthoryear{Ren et~al.}{2015}]{ren2015faster}
\begin{botherref}
\oauthor{\bsnm{Ren}, \binits{S.}},
\oauthor{\bsnm{He}, \binits{K.}},
\oauthor{\bsnm{Girshick}, \binits{R.}},
\oauthor{\bsnm{Sun}, \binits{J.}}:
Faster r-cnn: Towards real-time object detection with region proposal networks.
Advances in neural information processing systems
\textbf{28}
(2015)
\end{botherref}
\endbibitem

\bibitem[\protect\citeauthoryear{Krishna et~al.}{2017}]{krishna2017visual}
\begin{barticle}
\bauthor{\bsnm{Krishna}, \binits{R.}},
\bauthor{\bsnm{Zhu}, \binits{Y.}},
\bauthor{\bsnm{Groth}, \binits{O.}},
\bauthor{\bsnm{Johnson}, \binits{J.}},
\bauthor{\bsnm{Hata}, \binits{K.}},
\bauthor{\bsnm{Kravitz}, \binits{J.}},
\bauthor{\bsnm{Chen}, \binits{S.}},
\bauthor{\bsnm{Kalantidis}, \binits{Y.}},
\bauthor{\bsnm{Li}, \binits{L.-J.}},
\bauthor{\bsnm{Shamma}, \binits{D.A.}}, \betal:
\batitle{Visual genome: Connecting language and vision using crowdsourced dense image annotations}.
\bjtitle{International journal of computer vision}
\bvolume{123},
\bfpage{32}--\blpage{73}
(\byear{2017})
\end{barticle}
\endbibitem

\bibitem[\protect\citeauthoryear{He et~al.}{2017}]{he2017mask}
\begin{bchapter}
\bauthor{\bsnm{He}, \binits{K.}},
\bauthor{\bsnm{Gkioxari}, \binits{G.}},
\bauthor{\bsnm{Doll{\'a}r}, \binits{P.}},
\bauthor{\bsnm{Girshick}, \binits{R.}}:
\bctitle{Mask r-cnn}.
In: \bbtitle{Proceedings of the IEEE International Conference on Computer Vision},
pp. \bfpage{2961}--\blpage{2969}
(\byear{2017})
\end{bchapter}
\endbibitem

\bibitem[\protect\citeauthoryear{Gupta et~al.}{2019}]{gupta2019lvis}
\begin{bchapter}
\bauthor{\bsnm{Gupta}, \binits{A.}},
\bauthor{\bsnm{Dollar}, \binits{P.}},
\bauthor{\bsnm{Girshick}, \binits{R.}}:
\bctitle{Lvis: A dataset for large vocabulary instance segmentation}.
In: \bbtitle{Proceedings of the IEEE/CVF Conference on Computer Vision and Pattern Recognition},
pp. \bfpage{5356}--\blpage{5364}
(\byear{2019})
\end{bchapter}
\endbibitem

\bibitem[\protect\citeauthoryear{Jocher et~al.}{2022}]{glenn_jocher_2022_7347926}
\begin{botherref}
\oauthor{\bsnm{Jocher}, \binits{G.}},
\oauthor{\bsnm{Chaurasia}, \binits{A.}},
\oauthor{\bsnm{Stoken}, \binits{A.}},
\oauthor{\bsnm{Borovec}, \binits{J.}},
\oauthor{\bsnm{NanoCode012}},
\oauthor{\bsnm{Kwon}, \binits{Y.}},
\oauthor{\bsnm{Michael}, \binits{K.}},
\oauthor{\bsnm{TaoXie}},
\oauthor{\bsnm{Fang}, \binits{J.}},
\oauthor{\bsnm{imyhxy}},
\oauthor{\bsnm{Lorna}},
\oauthor{\bsnm{Yifu}, \binits{Z.}},
\oauthor{\bsnm{Wong}, \binits{C.}},
\oauthor{\bsnm{V}, \binits{A.}},
\oauthor{\bsnm{Montes}, \binits{D.}},
\oauthor{\bsnm{Wang}, \binits{Z.}},
\oauthor{\bsnm{Fati}, \binits{C.}},
\oauthor{\bsnm{Nadar}, \binits{J.}},
\oauthor{\bsnm{Laughing}},
\oauthor{\bsnm{UnglvKitDe}},
\oauthor{\bsnm{Sonck}, \binits{V.}},
\oauthor{\bsnm{tkianai}},
\oauthor{\bsnm{yxNONG}},
\oauthor{\bsnm{Skalski}, \binits{P.}},
\oauthor{\bsnm{Hogan}, \binits{A.}},
\oauthor{\bsnm{Nair}, \binits{D.}},
\oauthor{\bsnm{Strobel}, \binits{M.}},
\oauthor{\bsnm{Jain}, \binits{M.}}:
{ultralytics/yolov5: V7.0 - YOLOv5 SOTA Realtime Instance Segmentation}.
\doiurl{10.5281/zenodo.7347926} .
\url{https://doi.org/10.5281/zenodo.7347926}
\end{botherref}
\endbibitem

\bibitem[\protect\citeauthoryear{Shao et~al.}{2019}]{shao2019objects365}
\begin{bchapter}
\bauthor{\bsnm{Shao}, \binits{S.}},
\bauthor{\bsnm{Li}, \binits{Z.}},
\bauthor{\bsnm{Zhang}, \binits{T.}},
\bauthor{\bsnm{Peng}, \binits{C.}},
\bauthor{\bsnm{Yu}, \binits{G.}},
\bauthor{\bsnm{Zhang}, \binits{X.}},
\bauthor{\bsnm{Li}, \binits{J.}},
\bauthor{\bsnm{Sun}, \binits{J.}}:
\bctitle{Objects365: A large-scale, high-quality dataset for object detection}.
In: \bbtitle{Proceedings of the IEEE/CVF International Conference on Computer Vision},
pp. \bfpage{8430}--\blpage{8439}
(\byear{2019})
\end{bchapter}
\endbibitem

\bibitem[\protect\citeauthoryear{Kuznetsova et~al.}{2020}]{kuznetsova2020open}
\begin{barticle}
\bauthor{\bsnm{Kuznetsova}, \binits{A.}},
\bauthor{\bsnm{Rom}, \binits{H.}},
\bauthor{\bsnm{Alldrin}, \binits{N.}},
\bauthor{\bsnm{Uijlings}, \binits{J.}},
\bauthor{\bsnm{Krasin}, \binits{I.}},
\bauthor{\bsnm{Pont-Tuset}, \binits{J.}},
\bauthor{\bsnm{Kamali}, \binits{S.}},
\bauthor{\bsnm{Popov}, \binits{S.}},
\bauthor{\bsnm{Malloci}, \binits{M.}},
\bauthor{\bsnm{Kolesnikov}, \binits{A.}}, \betal:
\batitle{The open images dataset v4: Unified image classification, object detection, and visual relationship detection at scale}.
\bjtitle{International Journal of Computer Vision}
\bvolume{128}(\bissue{7}),
\bfpage{1956}--\blpage{1981}
(\byear{2020})
\end{barticle}
\endbibitem

\bibitem[\protect\citeauthoryear{Lin et~al.}{2014}]{lin2014microsoft}
\begin{bchapter}
\bauthor{\bsnm{Lin}, \binits{T.-Y.}},
\bauthor{\bsnm{Maire}, \binits{M.}},
\bauthor{\bsnm{Belongie}, \binits{S.}},
\bauthor{\bsnm{Hays}, \binits{J.}},
\bauthor{\bsnm{Perona}, \binits{P.}},
\bauthor{\bsnm{Ramanan}, \binits{D.}},
\bauthor{\bsnm{Doll{\'a}r}, \binits{P.}},
\bauthor{\bsnm{Zitnick}, \binits{C.L.}}:
\bctitle{Microsoft coco: Common objects in context}.
In: \bbtitle{Computer Vision--ECCV 2014: 13th European Conference, Zurich, Switzerland, September 6-12, 2014, Proceedings, Part V 13},
pp. \bfpage{740}--\blpage{755}
(\byear{2014}).
\bcomment{Springer}
\end{bchapter}
\endbibitem

\bibitem[\protect\citeauthoryear{Juneja et~al.}{2013}]{juneja2013blocks}
\begin{bchapter}
\bauthor{\bsnm{Juneja}, \binits{M.}},
\bauthor{\bsnm{Vedaldi}, \binits{A.}},
\bauthor{\bsnm{Jawahar}, \binits{C.}},
\bauthor{\bsnm{Zisserman}, \binits{A.}}:
\bctitle{Blocks that shout: Distinctive parts for scene classification}.
In: \bbtitle{Proceedings of the IEEE Conference on Computer Vision and Pattern Recognition},
pp. \bfpage{923}--\blpage{930}
(\byear{2013})
\end{bchapter}
\endbibitem

\bibitem[\protect\citeauthoryear{Lin et~al.}{2014}]{lin2014learning}
\begin{bchapter}
\bauthor{\bsnm{Lin}, \binits{D.}},
\bauthor{\bsnm{Lu}, \binits{C.}},
\bauthor{\bsnm{Liao}, \binits{R.}},
\bauthor{\bsnm{Jia}, \binits{J.}}:
\bctitle{Learning important spatial pooling regions for scene classification}.
In: \bbtitle{Proceedings of the IEEE Conference on Computer Vision and Pattern Recognition},
pp. \bfpage{3726}--\blpage{3733}
(\byear{2014})
\end{bchapter}
\endbibitem

\bibitem[\protect\citeauthoryear{Khan et~al.}{2016}]{khan2016discriminative}
\begin{barticle}
\bauthor{\bsnm{Khan}, \binits{S.H.}},
\bauthor{\bsnm{Hayat}, \binits{M.}},
\bauthor{\bsnm{Bennamoun}, \binits{M.}},
\bauthor{\bsnm{Togneri}, \binits{R.}},
\bauthor{\bsnm{Sohel}, \binits{F.A.}}:
\batitle{A discriminative representation of convolutional features for indoor scene recognition}.
\bjtitle{IEEE Transactions on Image Processing}
\bvolume{25}(\bissue{7}),
\bfpage{3372}--\blpage{3383}
(\byear{2016})
\end{barticle}
\endbibitem

\end{thebibliography}

\end{document}